\documentclass[runningheads]{llncs}
\usepackage{graphicx}
\usepackage{comment}
\usepackage{amsmath,amssymb} 
\usepackage{color,soul}
\usepackage{wrapfig}

\usepackage{bm}
\usepackage{booktabs}

\begin{document}
\pagestyle{headings}
\mainmatter
\def\ECCVSubNumber{5301}  

\title{Weakly-supervised Learning of Human Dynamics}

\titlerunning{Weakly-supervised Learning of Human Dynamics}
%
\author{Petrissa Zell\inst{1} \and
Bodo Rosenhahn\inst{1} \and
Bastian Wandt\inst{1}}
\authorrunning{Zell et al.}
%
\institute{Leibniz University Hannover, 30167 Hannover, Germany\\ 
\email{zell@tnt.uni-hannover.de}}
\maketitle

\begin{abstract}

This paper proposes a weakly-supervised learning framework for dynamics estimation from human motion. Although there are many solutions to capture pure human motion readily available, their data is not sufficient to analyze quality and efficiency of movements. Instead, the forces and moments driving human motion (\emph{the dynamics}) need to be considered. Since recording dynamics is a laborious task that requires expensive sensors and complex, time-consuming optimization, dynamics data sets are small compared to human motion data sets and are rarely made public. The proposed approach takes advantage of easily obtainable motion data which enables weakly-supervised learning on small dynamics sets and weakly-supervised domain transfer. Our method includes novel neural network (NN) layers for forward and inverse dynamics during end-to-end training. On this basis, a cyclic loss between pure motion data can be minimized, i.\,e.~no ground truth forces and moments are required during training.
The proposed method achieves state-of-the-art results in terms of ground reaction force, ground reaction moment and joint torque regression and is able to maintain good performance on substantially reduced sets.

\keywords{artificial neural networks, human motion, forward dynamics, inverse dynamics, weakly-supervised learning, domain transfer}
\end{abstract}

\section{Introduction}

\begin{figure}[ht]
    \centering
    \includegraphics[width=0.7\textwidth]{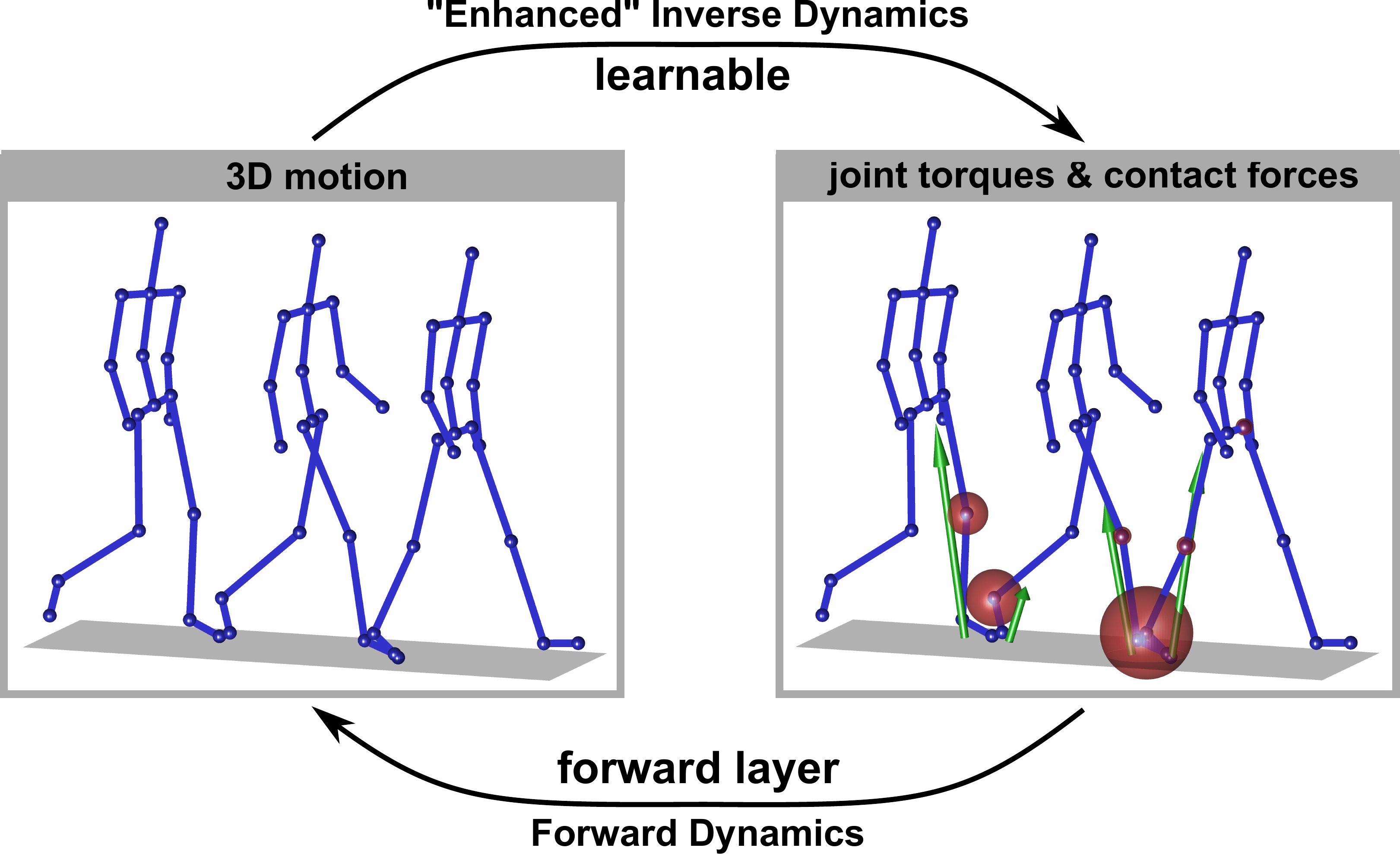}
    \caption{An \textit{Inverse Dynamics} method calculates joint torques based on an observed motion and contact forces. The enhanced task, that is implemented as a NN in this work, infers both, the joint torques and the exterior forces. The \textit{Forward Dynamics} step yields a simulated motion based on the acting forces and moments. With our forward layer this step can now be integrated into NN training.}
    \label{fig:teaser}
\end{figure}


Inverse dynamics describes the process of estimating net moments of force acting across skeletal joints from the three dimensional motion of the human skeleton and a set of exterior contact forces and moments. The obtained net moments are generally called joint torques (JT) and are of central interest in important fields, such as diagnostics of locomotor disorders, rehabilitation and prostheses design \cite{federolf13}, \cite{fregly07}, \cite{reinbolt09}. The JT cannot be measured non-intrusively and therefore need to be derived using computationally expensive optimization techniques. The common approaches are inverse and forward dynamics optimization. In inverse dynamics optimization, the kinematics of a model are optimized to match the target motion, while JT are inversely calculated based on the interaction with the ground and the model kinematics. For this step, 2nd order time derivatives are necessary, making the approach sensitive to noisy motion capture data. 

Forward dynamics optimization considers the reverse problem: The applied forces drive the motion of the human body. The equations of motion and the current motion state define an initial value problem, that can be solved by numerical integration, yielding a simulated motion. Included into an optimization framework, the forward dynamics step is utilized to find the optimal JT that generate a simulated motion with minimal distance to the captured motion. The necessary integration during each optimization step causes high computational complexity. Furthermore, both approaches, inverse and forward dynamics, require the measurement of exterior forces. In the case of locomotion, these exterior influences are the ground reaction forces and moments (GRF/M) acting at both feet. Hence, the analysis of human dynamics using conventional optimization techniques is restricted to a controlled laboratory setting with force plates to measure the GRF/M. To avoid the described problems and to achieve motion analysis in the wild, the exterior forces and the interior JT can be estimated directly from observed motions using machine learning techniques, e.\,g.~neural networks (NN).

Deep learning of human dynamics requires large data sets of human kinematics (3D motion of a kinematic model), exterior GRF/M and the driving JT. Unfortunately, corresponding data sets are few and often very restricted in terms of size and included motion types. To solve this problem, we propose a weakly-supervised deep learning method, that realizes NN training on small dynamics sets and domain transfer to new motion types, even without any ground truth of the acting forces. More precisely, our goal is to design a training process that is less depending on a large amount of dynamics data than a supervised baseline NN that learns to regress forces and moments from motion input. The proposed approach is called \textbf{Dynamics Network} and includes two novel dynamics layers: The forward (dynamics) layer and the inverse (dynamics) layer. The network replaces the traditional inverse dynamics task and enhances it by not only inferring the JT but also the GRF/M, as illustrated in Figure \ref{fig:teaser}. Subsequently, the forward layer solves the initial value problem given by the initial motion state and the network output. This results in a simulated motion that can be compared to the input motion, which gives rise to a loss purely defined on motion data. A full cycle, as illustrated in Figure~\ref{fig:teaser} is executed. For further control over the learned forces and moments we include the inverse layer, that penalizes GRF/M which do not match the observed accelerations. In contrast to the forward layer, it considers the ground reaction independently from JT, which allows for a decoupled control of both variables during training.

We demonstrate the benefits of our Dynamics Network by reducing the part of the used training sets, that include dynamical information, relying on the networks' own dynamics layers. Furthermore, we use our method to realize weakly- and unsupervised transfer learning between the related motion types, walking and running. We focus on locomotion since it represents the most important form of movement.

In summary, the contributions of this work are:
\begin{enumerate}
    \item A data set of 3D human kinematics with force plate measurements including various motion types.
    \item A novel forward dynamics layer that integrates the equations of motion and enables the minimization of a pure motion loss.
    \item A novel inverse dynamics layer that propagates forces and moments along the kinematic chain to measure the match between GRF/M and segment accelerations.
    \item Since the dynamics layers allow for training on pure motion samples without force information, we use this capacity in weakly-supervised learning and domain transfer between locomotion types.
\end{enumerate}

\section{Related work}
A laboratory setting with embedded force plates enforces strong restrictions on captured motions. Therefore, researchers increasingly exploit artificial NN to estimate ground reactions. The exterior forces are predicted based on 3D motions \cite{bastien19}, \cite{choi13}, \cite{johnson2018}, \cite{oh13}, accelerometer data \cite{leporace2015,leporace2018} or pressure insoles \cite{choi2018}, \cite{sim2015}. The seminal work by Oh et al.~\cite{oh13} includes a fully connected feed forward NN to solve the GRF ambiguity during double support. Based on the complete ground reaction information, the JT are inferred using a standard inverse dynamics method. An interesting approach proposed by Johnson et al.~\cite{johnson2018} encodes marker trajectories as RGB-images to make use of pre-trained CNNs for image classification.

Compared to GRF/M regression, relatively few works dealt with machine learning for the estimation of JT in human motion \cite{johnson14}, \cite{xiong19}. A method by Lv et al.~\cite{lv16} utilizes Gaussian mixture models to learn contact and torque priors that are included in a maximum a-posteriori framework. The priors help to lead the inverse dynamics optimization to realistic force and torque profiles. In a previous work \cite{zell19}, different machine learning algorithms are compared for the GRF/M and JT regression. The focus lies on contrasting end-to-end models to a hierarchical approach that subdivides the task into gait phase classification and regression.

It is worth mentioning, that machine learning for the inverse dynamics problem is also part of robotics research. Here, the common goal is to learn robotic control, such as JT, for trajectory planning. The latest works predominantly rely on recurrent NN and reinforcement learning \cite{devin17}, \cite{finn16}, \cite{mukhopadhyay19}, \cite{takahashi17}. 

In 2018, a differentiable physics engine for the inclusion into deep learning has been introduced and tested on simple simulated systems \cite{avila18}. A related work proposes a comparable physics engine and includes it in an RNN for learning of robotics control \cite{degrave16}. These models, however, are not sufficient to represent the human kinematic chain.
While robotics research focuses on the efficient execution of movement, the interaction with the environment and real-time application, the inverse dynamics problem for human motion is characterized by the complexity of the human locomotor system, which results in high data variability. Therefore, the focus primarily lies on the accurate estimation of forces and moments, in spite of the complex nature of the system, which makes regressors for human motion analysis highly dependent on large data sets.

The usage of cycle-consistency has already benefited visual correspondence tasks like temporal video alignment and cross-domain mapping of images, etc.~ \cite{Dwibedi_2019_CVPR}, \cite{hoffman2017cycada}, \cite{Shah_2019_CVPR}, \cite{Wang_2019_CVPR}, \cite{zhu2017}. The success of these methods has motivated us to apply cycle-consistency to human dynamics learning.

To the best of our knowledge, this work is the first to present differentiable NN layers for forward, as well as inverse dynamics of human motion. In contrast to all prior works, the proposed method can learn human dynamics in a weakly-supervised setting, without depending on complete dynamics ground truth. Our cycle consistent approach allows for the formulation of a pure motion loss, and thus drastically enlarges the usable data pool, even allowing for domain transfer between motion types.

\section{Human motion data sets}
\label{sec:datasets}

Deep learning of human dynamics demands data of the observed 3D motion, the acting GRF/M and the driving JT. We recorded 195 walking and 75 running sequences performed by 22 subjects of different gender and body proportions (demographic information can be found in Table \ref{tab:demographic}). The data was recorded using a Vicon motion capture system with synchronized AMTI force plates, embedded in the ground. An inverse kinematics algorithm was conducted to fit the skeleton of our physical model to the captured marker trajectories. In order to simplify the calculations of the forward layer, we apply a leg model with one torso segment to approximate the motion and the inertial properties of the whole upper body. This kind of model is typically used in locomotion analysis \cite{winter2009}. It's kinematics are completely represented by the generalized model coordinates $\bm q$ that consist of 6 global coordinates and 18 joint angles.

The GRF $\bm f_c$ and GRM $\bm m_c$ are calculated from the measured force plate data. Together they build the GRF/M vector
\begin{eqnarray}
    \bm F_{c} &=& \left[\bm f_{c_l}, \bm f_{c_r}, \bm m_{c_l}, \bm m_{c_r}\right]^T\,,\\
    \bm m_{c_i} &=& \bm d_{cop_i}\times \bm f_{c_i}+\bm t_{z_i}\,,\ \ i=l,r\,\,\nonumber
    \label{eq:Fc}
\end{eqnarray}
with the vector $\bm d_{cop}$ pointing from the foot center of mass to the center of pressure of the applied reaction force and the torsional torque $\bm t_z$. The index $i=l,r$ indicates the foot segment, where the respective part of $\bm F_c$ is applied.

Based on the kinematics and GRF/M we execute a forward dynamics optimization to receive the non-measurable JT $\bm \tau$. The applied forward dynamics step is equivalent to our forward layer, so that the resulting JT conform with the layer dynamics during training. We employ a sliding window approach for this pre-processing step, as well as for the network training and application. This way, the parameter space is decreased and the convergence is accelerated.

To further reduce the number of parameters, we apply polynomial fits to all relevant data types. This approximation also facilitates the learning of inverse dynamics, since the networks are not required to explicitly model temporal context. Without the polynomials, additional smoothness losses would be necessary to achieve continuous forces and moments during forward dynamics optimization, as well as NN training. The kinematics $\bm q$ and the GRF/M $\bm F_c$ are approximated by 3rd order polynomials. For the JT $\bm \tau$ we find, that linear approximations yield the best optimization results in a forward dynamics setting. The resulting polynomial coefficients are denoted by $\bm \gamma_q$, $\bm \gamma_f $ and $\bm \gamma_\tau$, respectively. Apart from these coefficients, each forward dynamics simulation is depending on the subject specific segment dimensions $\bm l_{sub}$. Together these parameters build a sample in our human motion data set:
\begin{equation}
    [\bm \gamma_q, \bm l_{sub}, \bm \gamma_f , \bm \gamma_\tau]\,.
\end{equation}
Using these approximations, we still achieve a representation of non-continuous contact by means of an overlapping window approach, that allows a discretization below window length.

Due to the restrictions introduced by the localized force plate measurements, approximately a third of the data set contains GRF/M. If the forward dynamics optimization converges to an insufficient minimum the corresponding torques cannot be included, so that even a smaller part of the data includes JT. We divide our data set accordingly into a pure motion subset, a motion and GRF/M subset and a subset with the complete data. These subsets are referred to as \textit{motion-set}, \textit{ground-reaction-set} and \textit{torque-set}, respectively. To investigate a further scenario, that requires less supervision and measurements, we define a \textit{contact-set}, which contains motion states and binary information about the ground contact, i.e.~which foot is in contact with the ground. The presented data sets are used in different training modes that represent various levels of supervision. 

\section{Dynamics Network}
\label{sec:net}

\begin{figure}
    \centering
    \includegraphics[width=0.95\textwidth]{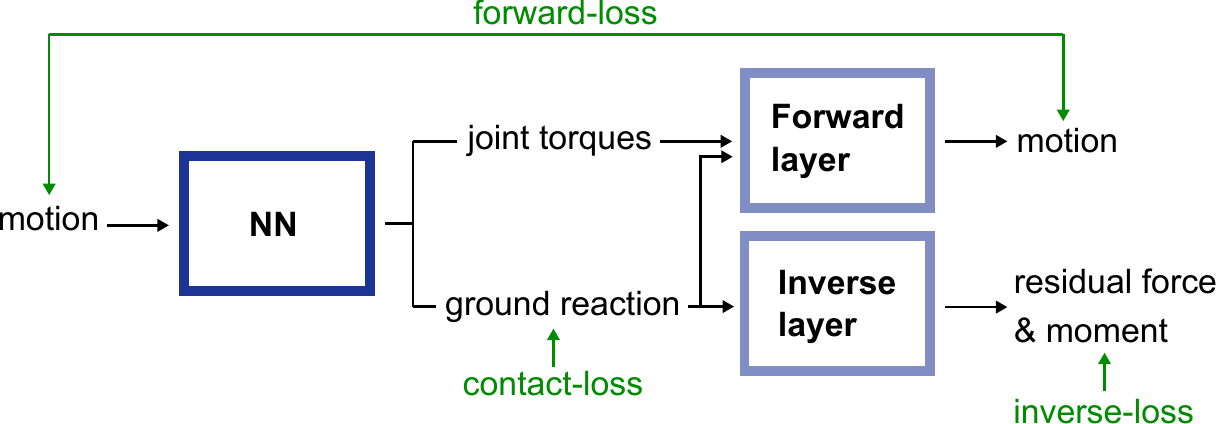}
    \caption{Schematic structure of the Dynamics Network. The output of the NN is processed using forward and inverse layer to accomplish training without GRF/M and JT ground truth.}
    \label{fig:net}
\end{figure}

The structure of our Dynamics Network is presented in Figure \ref{fig:net}. A fully connected NN executes the extended inverse dynamics task from motion to JT and GRF/M. The network has a bottle-neck structure with 5 fully connected layers, containing about 5800 parameters in total and Leaky-ReLu activations \cite{Maas2013RectifierNI}. The forward dynamics step is implemented as a NN layer (forward layer) and yields simulated motion states based on the network output. A detailed description is given in Section \ref{sec:fd}. Combined, the NN and the forward layer build a cycle that enables the minimization of a loss between motion states. The additional inverse layer measures the consistency of the input motion with the predicted GRF/M by propagating forces and moments along the kinematic chain and calculating the residuals at the last segment. A complete description follows in Section \ref{sec:id}.
Furthermore, the network can be trained in a supervised manner, using a mean squared error (MSE) on the predicted GRF/M and JT. This approach will be used as a baseline. Since the corresponding ground truth data is not always available, our dynamics layers can be used to simulate and control it in a weakly-supervised setting. To gradually reduce the level of supervision, we implement a contact-loss, that penalizes forces during time frames with no ground contact. It only requires binary information.

Our methods operate with four different loss functions, that can be activated separately or in combination. On this basis, we define and compare different training modes that realize various levels of supervision and operate on different subsets of the data: Starting with full supervision, the baseline method uses the MSE of the predicted parameters $\gamma_f^\mathrm{pred}$ and $\gamma_\tau^\mathrm{pred}$ for GRF/M and JT, respectively:
\begin{equation}
    L_\mathrm{mse} = \|\bm \gamma_f^\mathrm{pred} - \bm\gamma_f^\mathrm{true}\|^2_2 + \|\bm \gamma_\tau^\mathrm{pred} - \bm \gamma_\tau^\mathrm{true}\|^2_2\,.
    \label{eq:L_mse}
\end{equation}
The baseline network is exclusively trained on the torque-set and the ground-reaction-set.

In the next training mode, the network is additionally trained using the contact-set with a reduced level of supervision. We assume, that only binary information about the contact state is available in this subset. The corresponding training mode is referred to as \textit{contact-forward-inverse-training} or in short \textit{cFI-training}. During this mode, the corresponding network (\textit{cFI-net}) is trained in an alternating procedure: If data from the contact-set is chosen as input, the procedure switches between the contact-loss, the inverse-loss and the forward-loss. Based on Eq.\,(\ref{eq:Fc}), we define the contact-loss as
\begin{equation}
    L_\mathrm{contact} = \|\bm F_{c_i}(c_i=0)\|^2_2\,,
\end{equation}
with the contact state $c_i=1$ if foot $i$ is in contact with the ground and $c_i=0$, otherwise. If data from the torque-set or the ground-reaction-set is chosen, the MSE of Eq.~(\ref{eq:L_mse}) is minimized, similarly to the baseline method.

The third training mode is termed \textit{forward-training} or \textit{F-training}. In addition to supervised training on the torque-set and the ground-reaction-set, the network, \textit{F-net}, is trained in an unsupervised setting, using the motion-set and minimizing the forward-loss.

In all training modes the included losses are minimized alternatingly. To balance the influence of different loss functions, we use adaptive weighting, that is updated after each epoch according to the ratio of the observed average values during the last epoch.

In Section \ref{sec:experiments} we compare the proposed training modes regarding their performance on small data sets and their capability to perform domain transfer from walking to running motions.

\subsection{Forward dynamics layer}
\label{sec:fd}
In this section, we describe our forward dynamics simulation and the implementation as a NN layer. We choose a simple model and a basic numerical integration technique in order to maintain relatively low computational complexity. This is necessary to facilitate the integration in NN training.

Our human body model is a leg model with one additional segment that presents the mean upper body kinematics. As mentioned before, the kinematic state of the model is fully represented by the generalized coordinates $\bm q$ and their derivatives $\bm{\dot{q}}$. The generalized coordinates include 18 joint degrees of freedom that are effected by the JT $\bm \tau$. The GRF/M $\bm F_c$ introduced in Eq.~(\ref{eq:Fc}) is applied at the center of mass of the respective foot segment. Each segment is associated with a mass and a tensor of inertia that are approximated from the segment shape and literature values of the population \cite{winter2009}. For this step, we model the segments as simple geometric shapes with constant density. The segment volume is scaled according to the length based on average scaling factors of the data set. With this assumption, the segment lengths $\bm l_{sub}$ completely describe the body model.

We formulate the equation of motion using the TMT-method \cite{schwab09} similar to \cite{brubaker08}, \cite{zell_chapter}. The resulting 2nd order differential equation has the following form:
\begin{equation}
    \bm{\mathcal{M}}(\bm q_t, \bm l_{sub})\ddot{\bm q}_t = \bm{\mathcal{F}}(\bm q_t, \dot{\bm q}_t, \bm F_{c_t}, \bm\tau_t, \bm l_{sub})
\end{equation}
Here, $\bm{\mathcal{M}}$ is the generalized inertia matrix and the right hand side is the sum of all acting forces. For better readability we drop the time frame index $t$ in the remaining part of this work.
The problem is reformulated as a 1st order differential equation by introducing the state vector $\bm x=[\bm q^T, \bm{\dot q}^T]^T$:
\begin{equation}
    \bm{\dot x} = \begin{bmatrix} \bm{\dot q}\\
                                \bm{\mathcal{M}}(\bm q, \bm l_{sub})^{-1}\bm{\mathcal{F}}(\bm x, \bm F_c, \bm\tau, \bm l_{sub})
    \end{bmatrix}
    \label{eq:EOM}
\end{equation}
Together with an initial state vector $\bm x_0$ this equation gives rise to an initial value problem, that can be solved by numerical integration.

During integration an acceleration error propagates with the squared integration time. The same applies to an error in JT and GRF/M. In order to reduce this high sensitivity for NN training, we propose to apply a damping factor to $\bm{\dot x}$. The damping is based on the standard deviation $\bm\sigma_{\dot x}$ of the absolute velocities and accelerations found in the training set and on the maximum absolute values $\bm m_{\dot x}$ that occur in the current sample. For each component $j$ the damping is set to
\begin{equation}
    d_j = \mathrm{exp}\left\{-\mathrm{max}\left(\frac{|\dot x_{j}|-m_{\dot x,j}-k\sigma_{\dot x,j}}{k\sigma_{\dot x,j}}, 0\right)\right\}\,,
    \label{eq:damping}
\end{equation}
where $\dot x_j$ is the undamped component of the current integration step. The resulting vector $\bm d$ is included by building the Hadamard product to result in the damped equation of motion:
\begin{equation}
    \bm{\dot x} = \begin{bmatrix} \bm{\dot q}\\
                                \bm{\mathcal{M}}^{-1}\bm{\mathcal{F}}
    \end{bmatrix}\odot \bm d\,.
    \label{eq:EOMdamp}
\end{equation}
The parameter $k$ in Eq.~(\ref{eq:damping}) determines the steepness of the damping curve and the starting point of the decrease from one. In simple terms, it broadens the region of acceptable accelerations to $k$ times of the standard deviation. The value is heuristically set to $k=10$. We found, that this setting results in stable simulations that can still be optimized during training.

For numerical integration we apply Euler's method with constant step size to keep the computation time as small as possible. Consequently, our forward layer can be seen as a function $FD$, that executes $n=(\text{window size} -1)$ Euler steps. It receives the input parameters 
\begin{equation}
    \bm p = (\bm x_0, \bm l_{sub}, \bm F_{c_{1\hdots n}}, \bm\tau_{1\hdots n}, \bm m_{\dot x})
\end{equation}
and outputs the simulated motion states 
\begin{equation}
    \bm x^{sim}_{1\hdots n} = FD(\bm p)\,.
\end{equation}
Based on this result we define the forward loss as
\begin{equation}
    L_\mathrm{forward}=\mathrm{MSE}(\bm x^{sim}_{1\hdots n} , \bm x^{true}_{1\hdots n} ) + \alpha \|\bm d - \bm 1\|_1\,,
    \label{eq:sim_loss}
\end{equation}
with weighting factor $\alpha=1$ and an $L_1$-loss to penalize damping factors smaller than one.

For the back propagation of $L_\mathrm{forward}$ during NN training, the gradients of the output states with respect to the input of the layer need to be known. This can either be achieved using automatic differentiation included in most deep learning frameworks or by explicit calculation using sensitivity analysis. The latter approach is described in Appendix \ref{app:grad}.

\subsection{Inverse dynamics layer}
\label{sec:id}
The inverse layer receives the ground truth motion and the predicted GRF/M as input and propagates forces and moments along the kinematic chain in a bottom-up procedure. The calculation starts at the centers of mass of the model's feet, where GRF/M are applied. Each segment is considered in a free body diagram to deduce the forces and moments at the proximal joint based on the acting forces and moments at the distal joint and the linear and angular acceleration of the segment. For segment $s$ the force $\bm F_p$ effecting the proximal joint is given by
\begin{equation}
    \bm F_p = m_s(\bm a_s-\bm g) - \bm F_d\,,
\end{equation}
with the distal force $\bm F_d$, the segment mass $m_s$, the segment acceleration $\bm a_s$ and the gravitational acceleration $\bm g$.
The moment $\bm M_p$ acting across the proximal joint can be calculated in a related manner:
\begin{equation}
    \bm M_p = \bm I_s \bm \alpha_s - \bm M_d - \sum_{j=p,d}\bm r_j \times \bm F_j\,,
\end{equation}
where $\bm M_d$ denotes the distal moment, $\bm I_s$ is the tensor of inertia for the considered segment and $\bm \alpha_s$ is its angular acceleration. The cross products account for moments resulting from the linear forces applied at the joints with $\bm r_j$ being the vector from segment center of mass to the joint coordinates.

Based on these equations, the forces and moments are propagated along the kinematic chain, resulting in a residual force $\bm F_\mathrm{res}$ and a residual moment $\bm M_\mathrm{res}$ at the end of the chain, in our case, the center of mass of the upper body. If the model accelerations perfectly match the GRF/M these residuals are equal to zero, so that the inverse loss is defined as
\begin{equation}
    L_{\mathrm{inverse}} = \|\bm F_\mathrm{res}\|^2_2 + \|\bm M_\mathrm{res}\|^2_2\,.
\end{equation}
The gradient calculation for the back propagation through the inverse layer is described in Appendix \ref{app:grad}.

\section{Experiments}
\label{sec:experiments}
In this section, the proposed methods are evaluated regarding their capability to learn exterior GRF/M and interior JT from motion. In particular, weakly-supervised learning on small training sets and domain transfer are investigated.

In the following experiments the root mean squared error (RMSE) and the relative RMSE (rRMSE), denoted by $\epsilon$ and $\epsilon_r$, respectively, are used to quantitatively evaluate regression results. The rRMSE is normalized to average value ranges of the training set:
\begin{equation}
    \epsilon_r = \frac{\epsilon}{\frac{1}{N}\sum_{i\in \text{train set}}[\max(v_t)-\min(v_t)]}\,.
\end{equation}
Here, $u$ is the predicted variable and $v$ is the target variable. The number of training samples is denoted by $N$.
All experiments are executed with three random splits into training, validation and test set. The splits are done subject wise, meaning, that sequences of the same subject are exclusively included in one set.

In addition to the experiments, presented in the following sections, an exemplary application to CMU data \cite{cmumocap} and a noise experiment are included in the appendix.


\subsection{Comparison to the state of the art}
\label{sec:sota}
In a first experiment, we compare the proposed Dynamics Network and our baseline network to state-of-the-art methods for the inference of GRF/M together with JT. For this purpose, we use the gait set including slow and fast walking. Table \ref{tab:exp0} lists the corresponding results. 

\begin{table*}[]
    \centering
    \caption{RMSE $\epsilon$ and rRMSE $\epsilon_r$ of GRF/M and JT regression results of the gait data set.}
    \begin{tabular}{lccccc}
    \toprule
    method& \ $\epsilon_f$ [N/kg]\ & \ $\epsilon_{r_f}$ [\%]\ & \ $\epsilon_m$ [Nm/kg]\ & \ $\epsilon_{r_m}$ [\%]\ & \ $\epsilon_\tau$ [Nm/kg]\\
    \midrule
    Lv et al.~\cite{lv16} & 0.700 & 20.3 & 0.077 & 28.1  & - \\
    Zell et al.~\cite{zell19} & \textbf{0.388} & \textbf{13.1} & \textbf{0.041} & \textbf{21.1} & 0.055 \\
    Baseline net & 0.591 & 14.4 & 0.056 & 21.2 & 0.055 \\
    F-net & 0.626 & 14.9 & 0.059 & 22.1 & \textbf{0.053} \\
    cFI-net & 0.733 & 14.7 & 0.064 & 22.5 & 0.054 \\
    \bottomrule
    \end{tabular}
    \label{tab:exp0}
\end{table*}


While GRF/M ground truth can be calculated directly from force plate measurements and skeleton fits, the JT ground truth is a dynamics optimization result, associated with a higher uncertainty. Especially using forward dynamics, there are failure cases when the optimization does not converge to a satisfactory minimum. A comparison to these optimized sequences is not informative. Therefore, the JT evaluation is performed sample wise and only for the learning-based approaches to allow for a fair comparison. The GRF/M are evaluated as complete sequences. 

The best performing method with regards to GRF/M regression \cite{zell19}, uses an SVM for gait phase classification and then regresses force and moment parameters on the resulting class subsets via Random Forests. In contrast to this approach the baseline and the Dynamics Network can be trained end-to-end and yield competitive results. Regarding the JT results, the Dynamics Network slightly outperforms the other methods. Instead of minimizing a loss on the JT data directly, the forward loss regards the impact, the JT have on a simulated motion, which results in a stronger loss function. The method by Lv et al.~\cite{lv16} is a maximum a-posteriori inverse dynamics optimization, \begin{wrapfigure}[14]{r}{0.5\linewidth}
    \centering
    \includegraphics[width=0.4\textwidth]{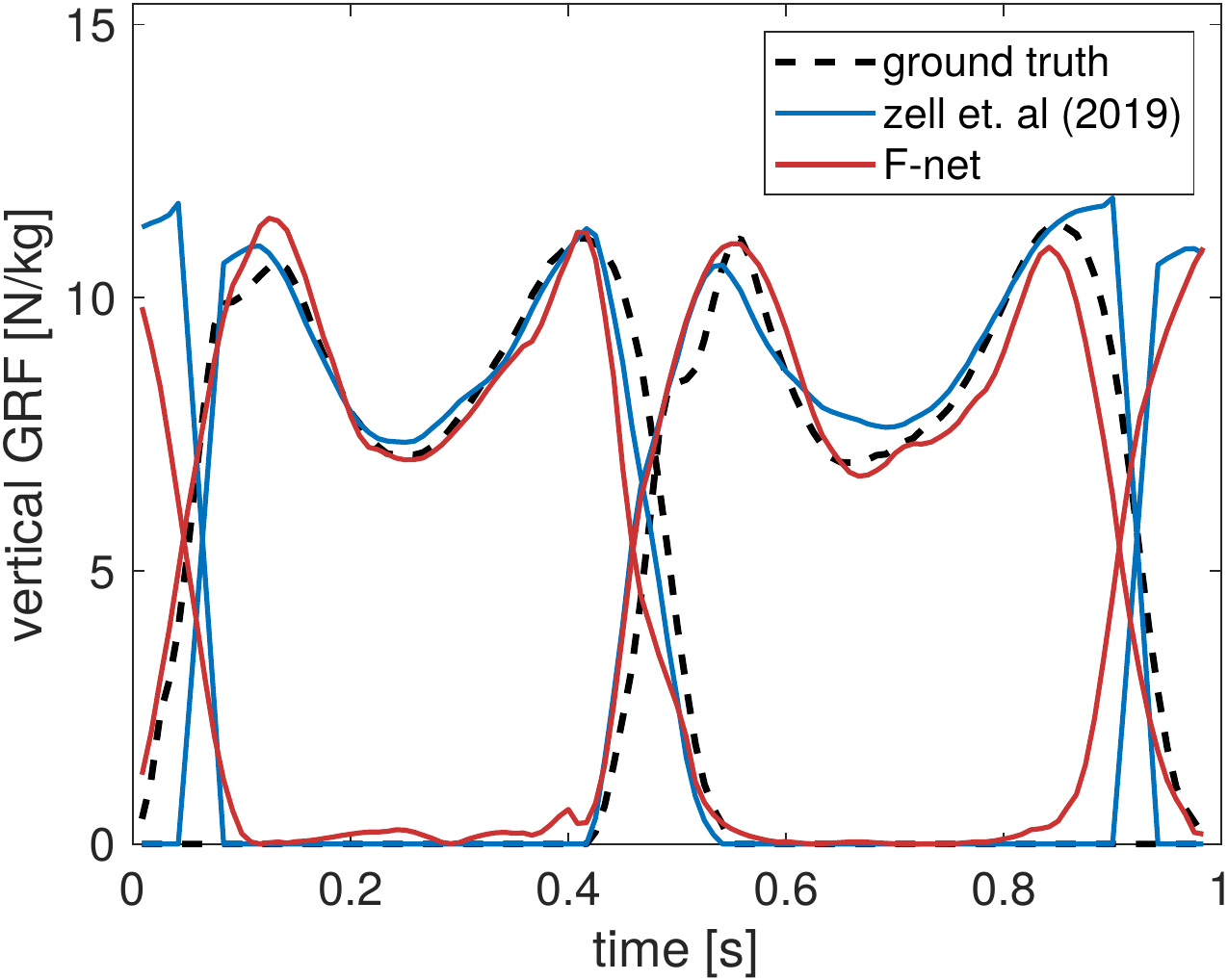}
    \caption{Comparison of predicted vertical GRF by \cite{zell19} and F-net}
    \label{fig:compare}
\end{wrapfigure}which incorporates a data-prior to guide the optimization towards realistic forces. The values were generated using our own implementation of the approach.

A qualitative comparison between F-net and \cite{zell19}, depicted in Figure \ref{fig:compare}, shows that although \cite{zell19} achieves low error values, the primary gait phase classification may lead to rapidly changing forces at gait phase transitions. This physically implausible behaviour is inherent to the hierarchical structure of the method and does not occur using the proposed network. 

\subsection{Learning dynamics on small data sets}
\label{sec:small_sets}
The first goal of this work is to learn human locomotion dynamics on small data sets without overfitting to the training samples. In Section \ref{sec:net}, we introduced two training modes with different degree of supervision to achieve this goal. These training modes are now compared to each other regarding their capability to operate on a training set with decreasing numbers of ground-reaction-set and torque-set samples. The motion-set, that only contains motion information, is always included to its full extend, so that the networks trained with forward and inverse layer, respectively, receive data of all subjects (contained in the training set). This way, we can show the effect of the higher data variability and the benefit of our dynamics layers. For cFI-training the contact-set is used, as well. 

\begin{figure}
    \begin{minipage}{0.49\textwidth}
        \centering
        \includegraphics[width=0.85\textwidth]{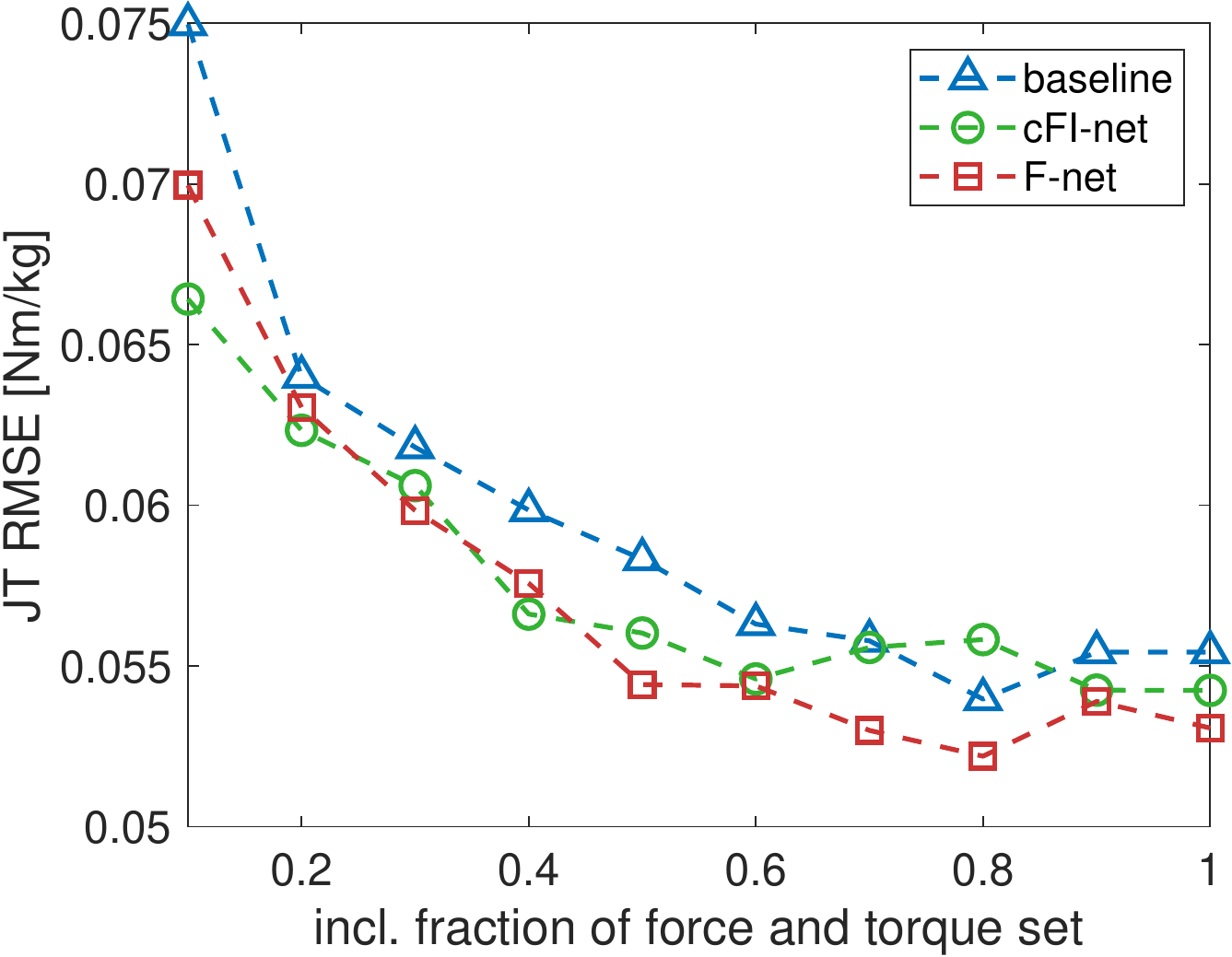}
    \end{minipage}
    \begin{minipage}{0.49\textwidth}
        \centering
        \includegraphics[width=0.85\textwidth]{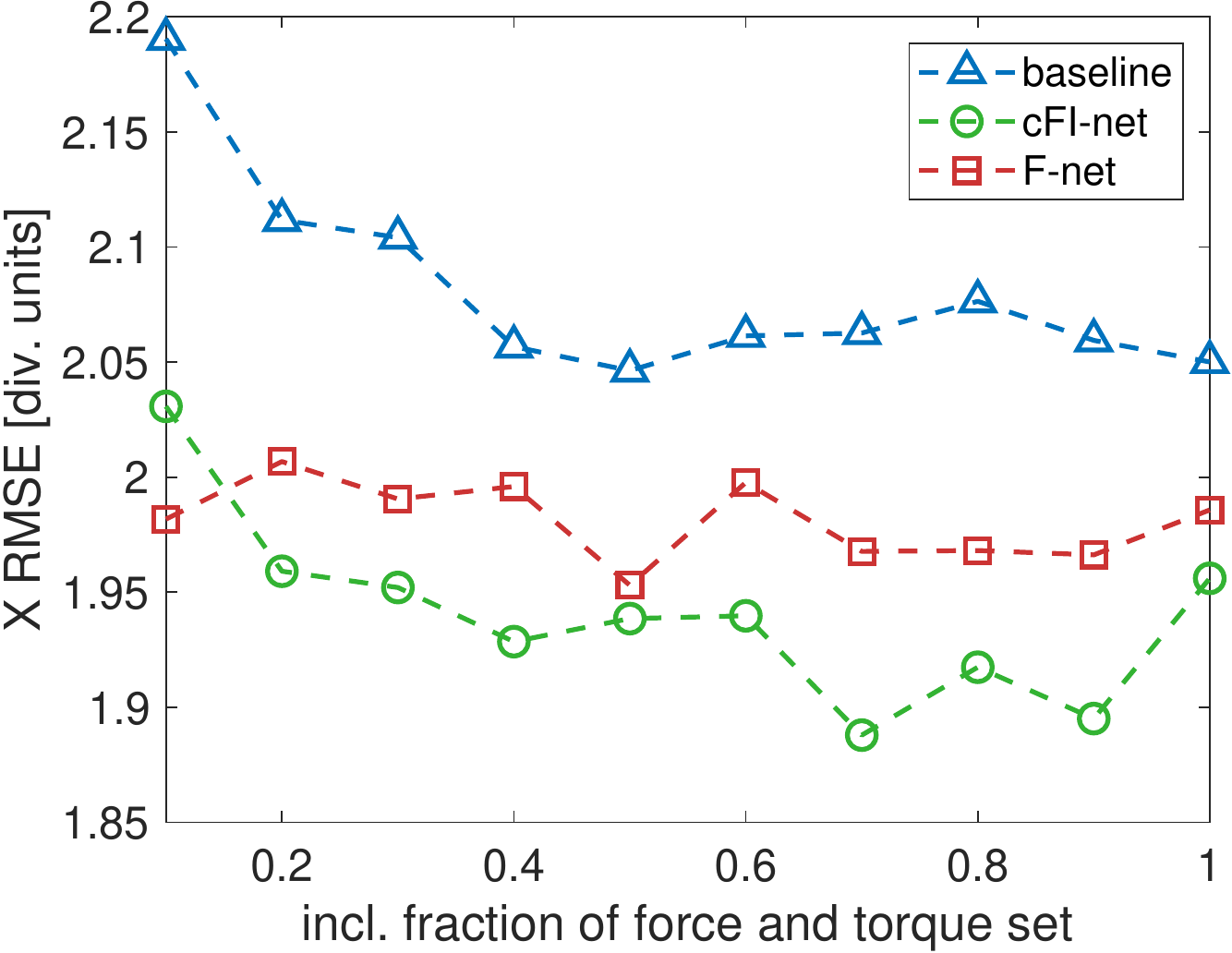}
    \end{minipage}
    \caption{Reduction of ground reaction and torque samples included in the training set. The left side shows RMSE of JT and the right side the associated RMSE of simulated motion states.}
    \label{fig:fracTau}
\end{figure}
\begin{figure}
    \begin{minipage}{0.327\textwidth}
        \centering
        \includegraphics[width=0.95\textwidth]{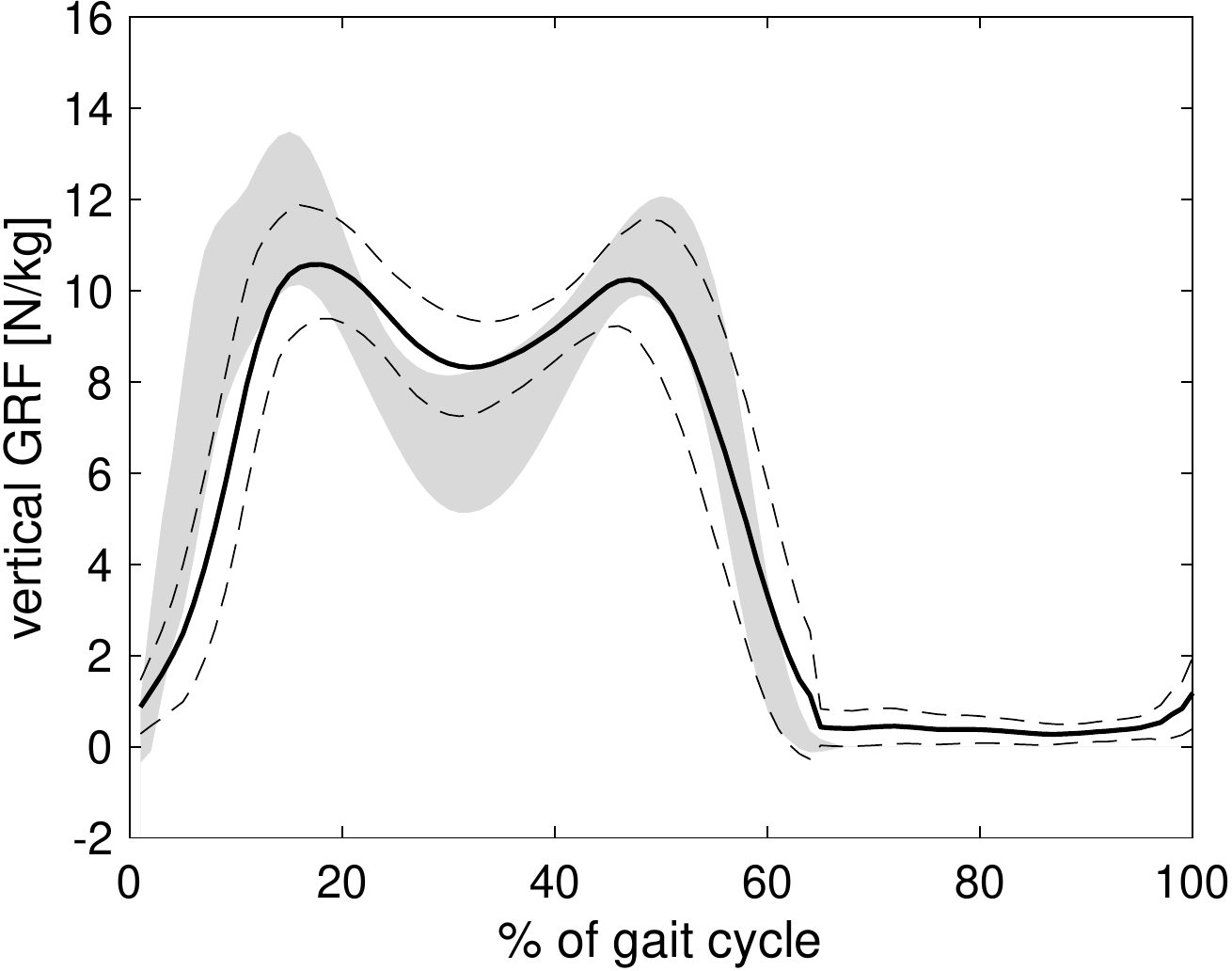}
    \end{minipage}
    \begin{minipage}{0.327\textwidth}
        \centering
        \includegraphics[width=0.95\textwidth]{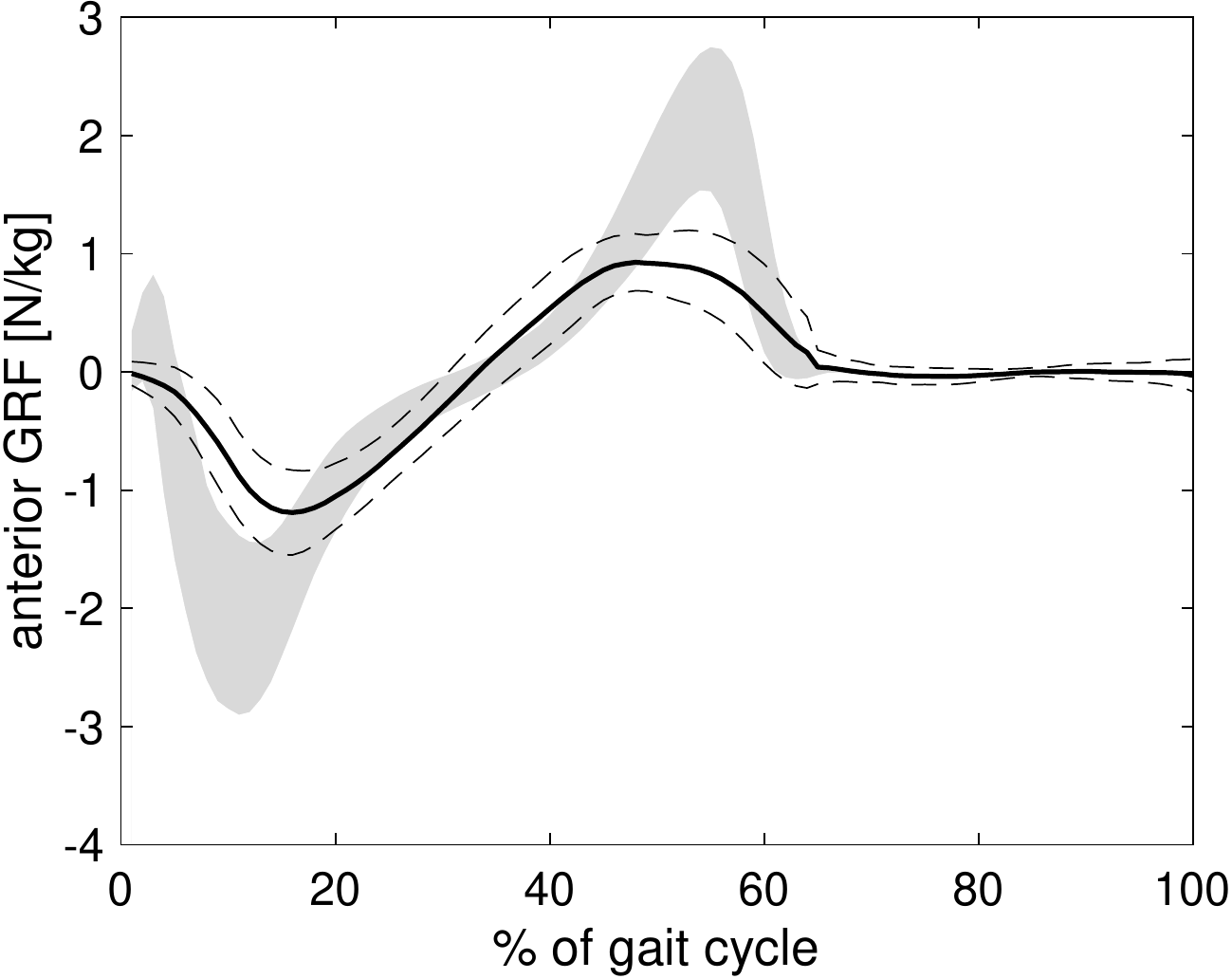}
    \end{minipage}
    \begin{minipage}{0.327\textwidth}
        \centering
        \includegraphics[width=0.95\textwidth]{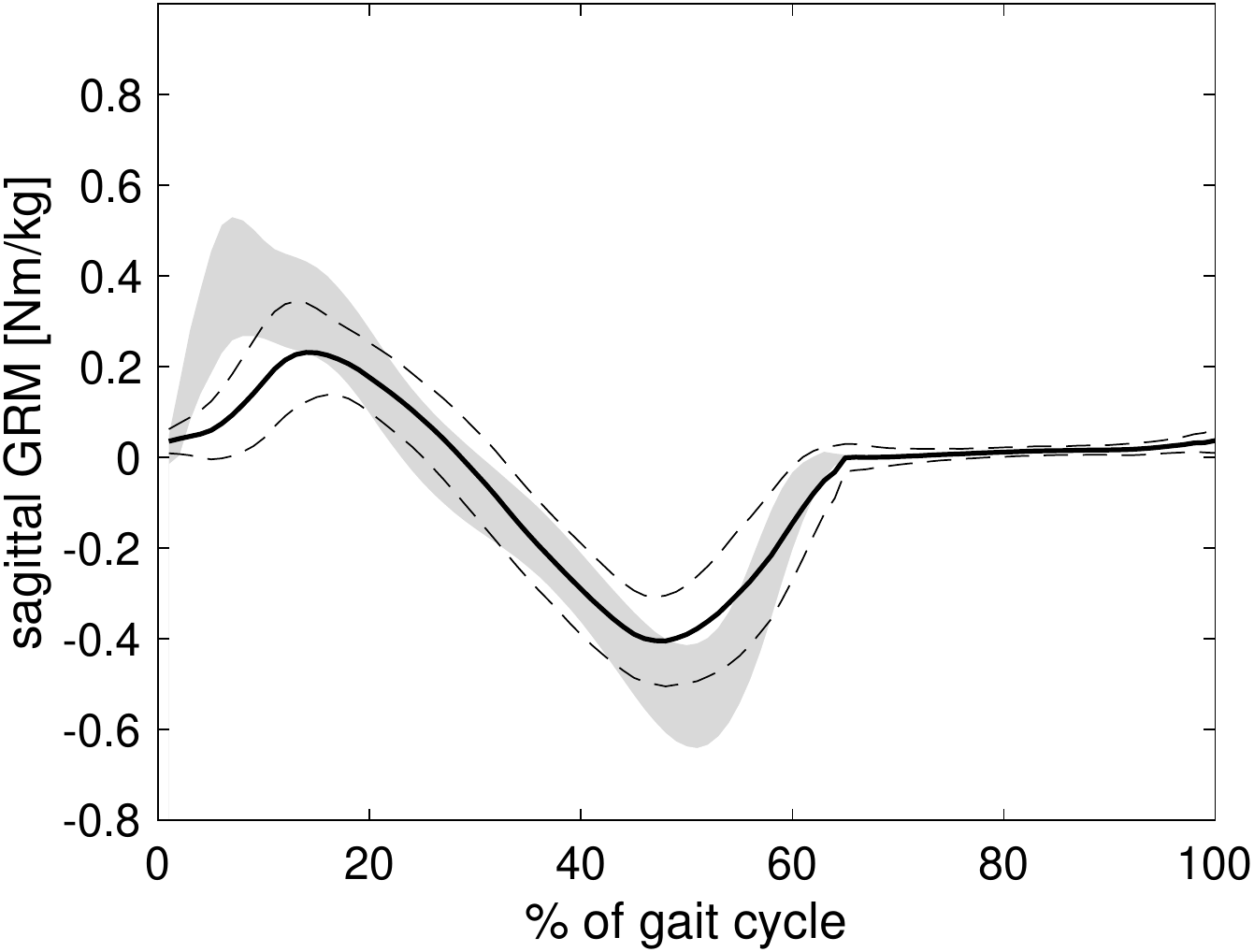}
    \end{minipage}
    \begin{minipage}{0.327\textwidth}
        \centering
        \includegraphics[width=0.95\textwidth]{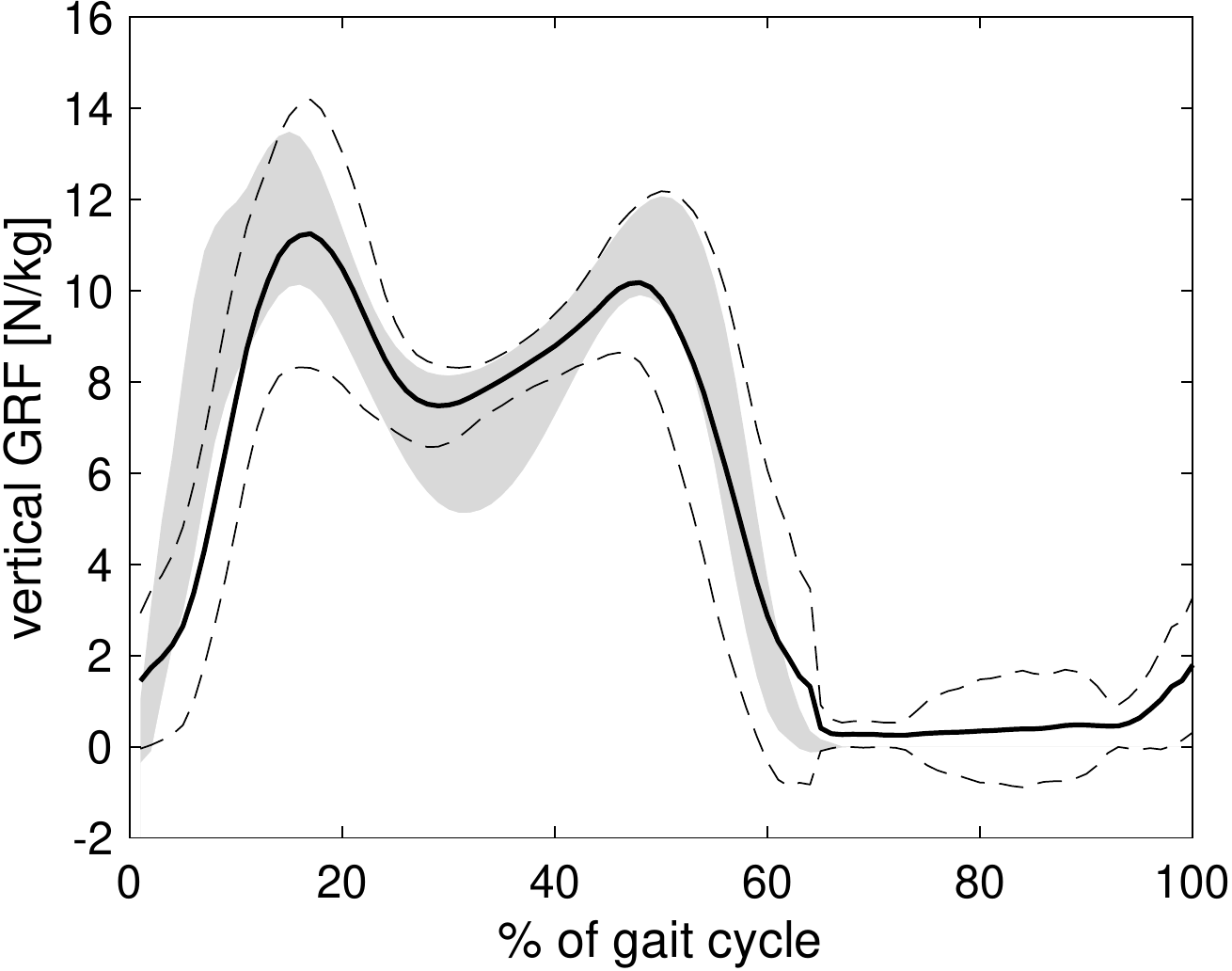}
    \end{minipage}
    \begin{minipage}{0.327\textwidth}
        \centering
        \includegraphics[width=0.95\textwidth]{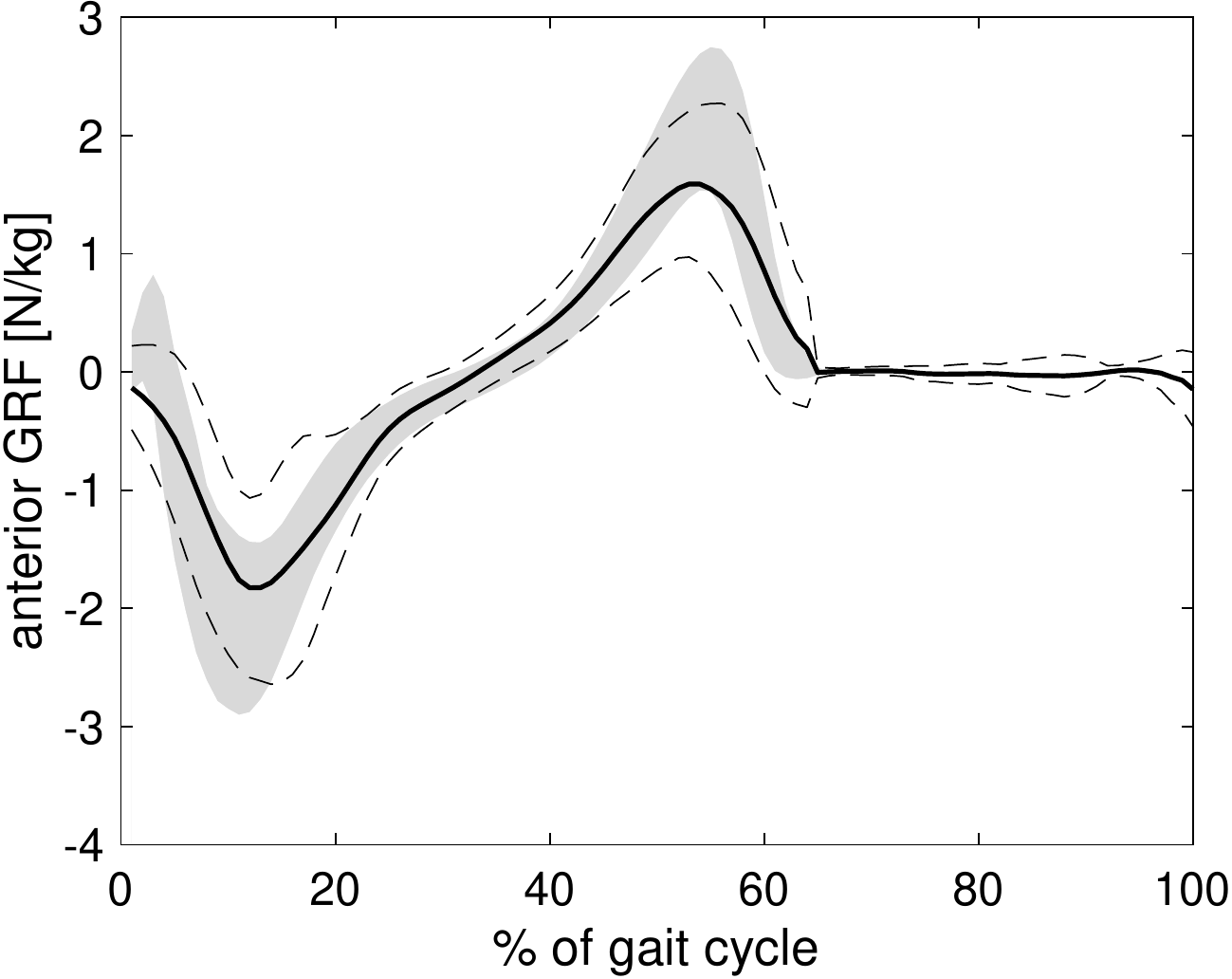}
    \end{minipage}
    \begin{minipage}{0.327\textwidth}
        \centering
        \includegraphics[width=0.95\textwidth]{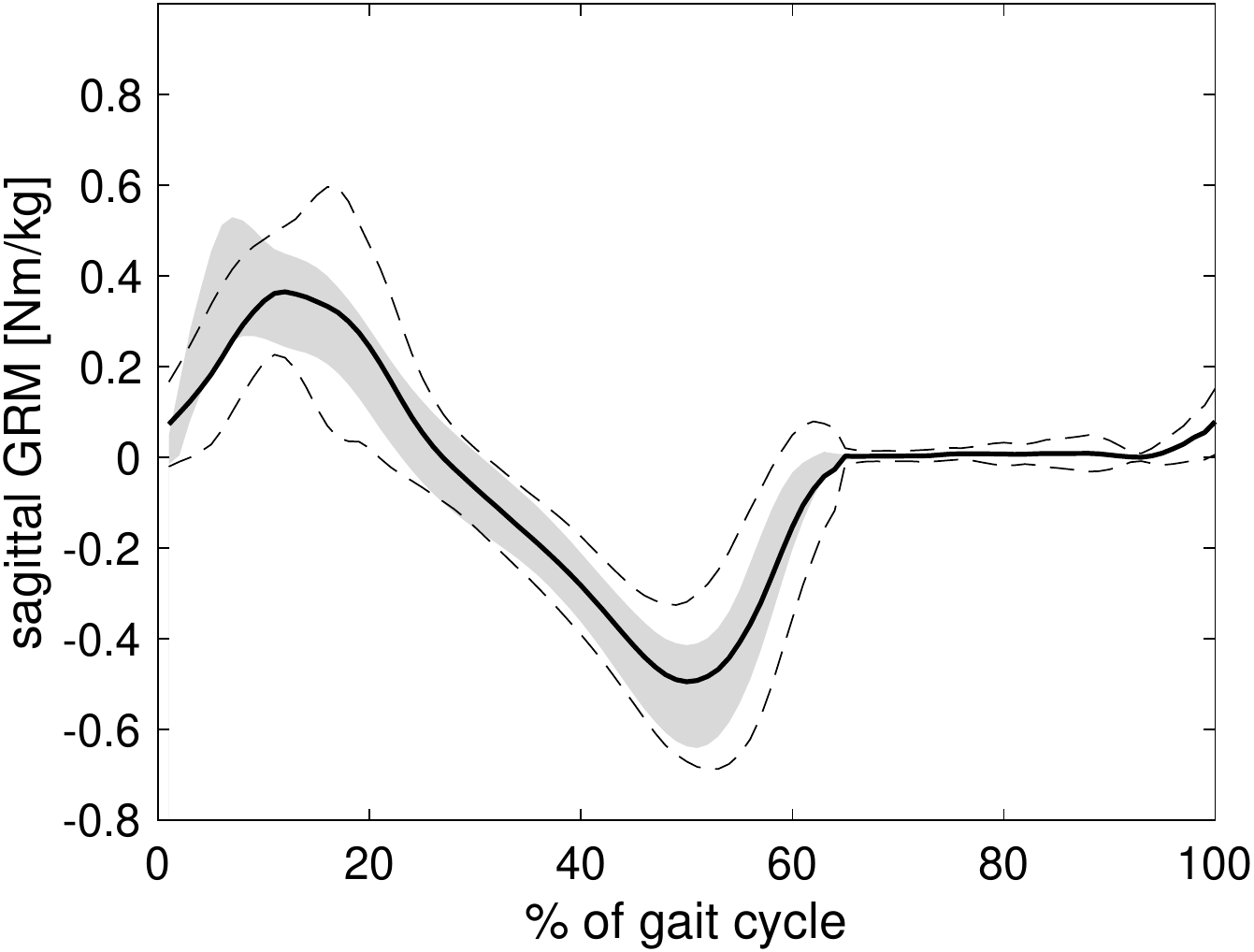}
    \end{minipage}
    \begin{minipage}{0.327\textwidth}
        \centering
        \includegraphics[width=0.95\textwidth]{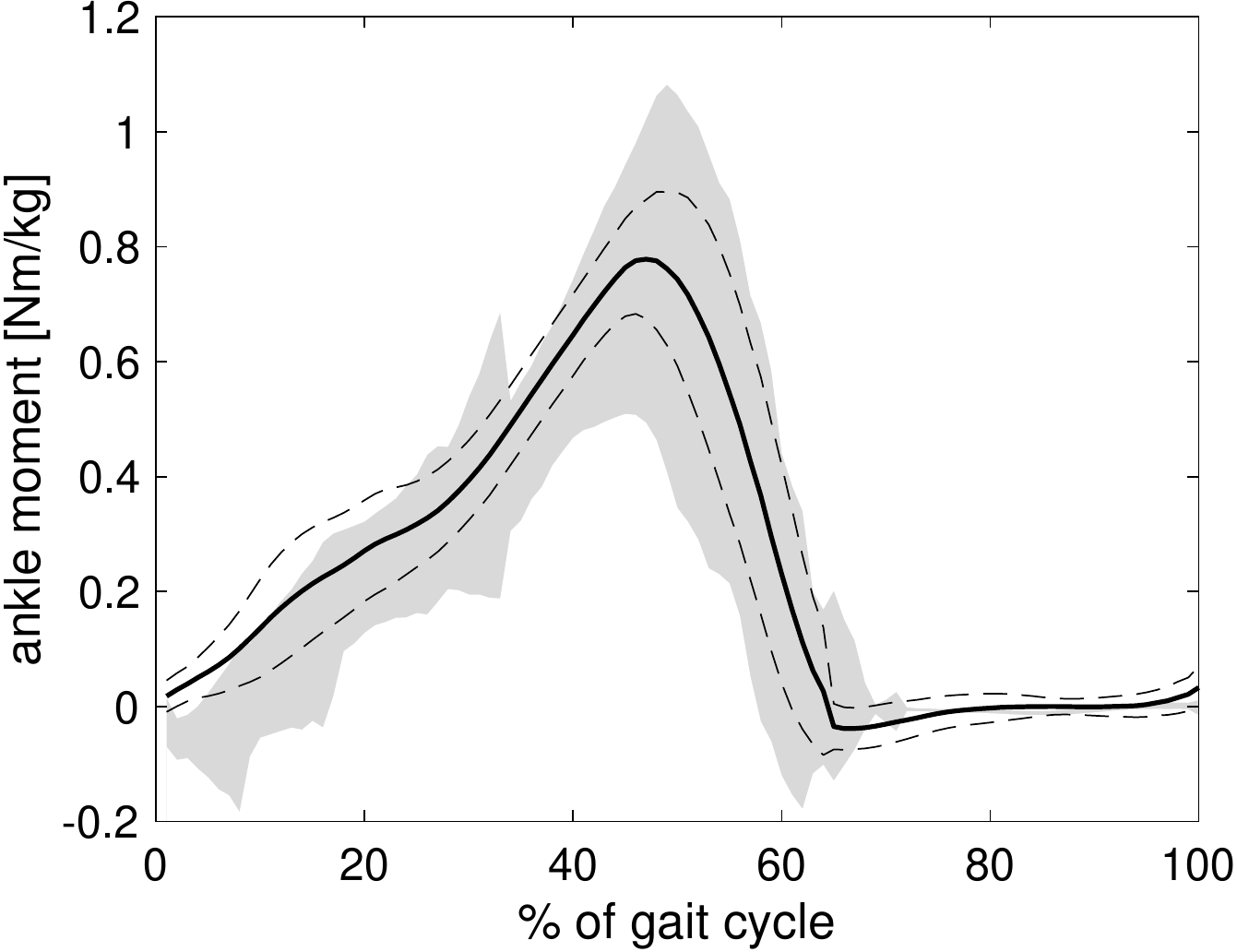}
    \end{minipage}
    \begin{minipage}{0.327\textwidth}
        \centering
        \includegraphics[width=0.95\textwidth]{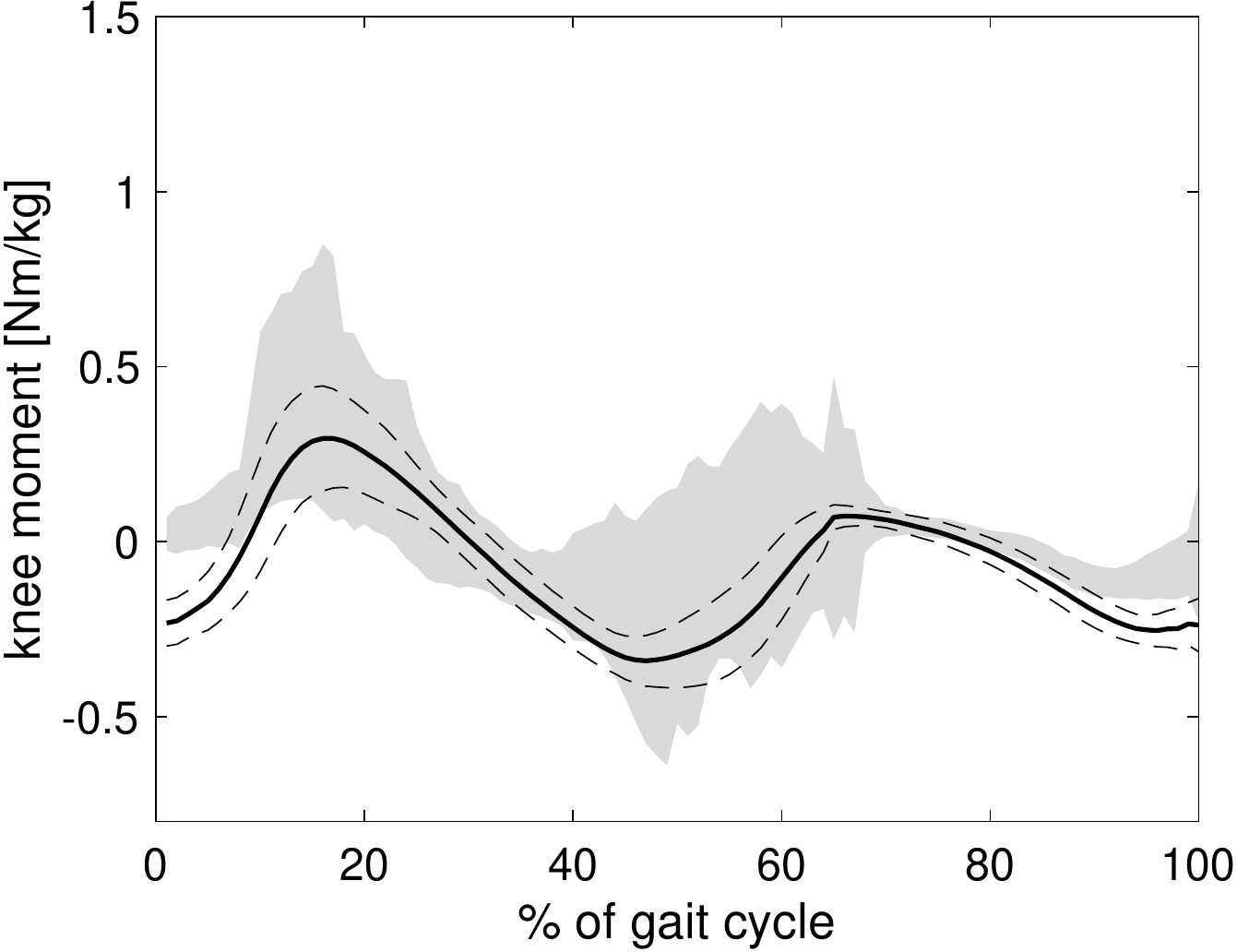}
    \end{minipage}
    \begin{minipage}{0.327\textwidth}
        \centering
        \includegraphics[width=0.95\textwidth]{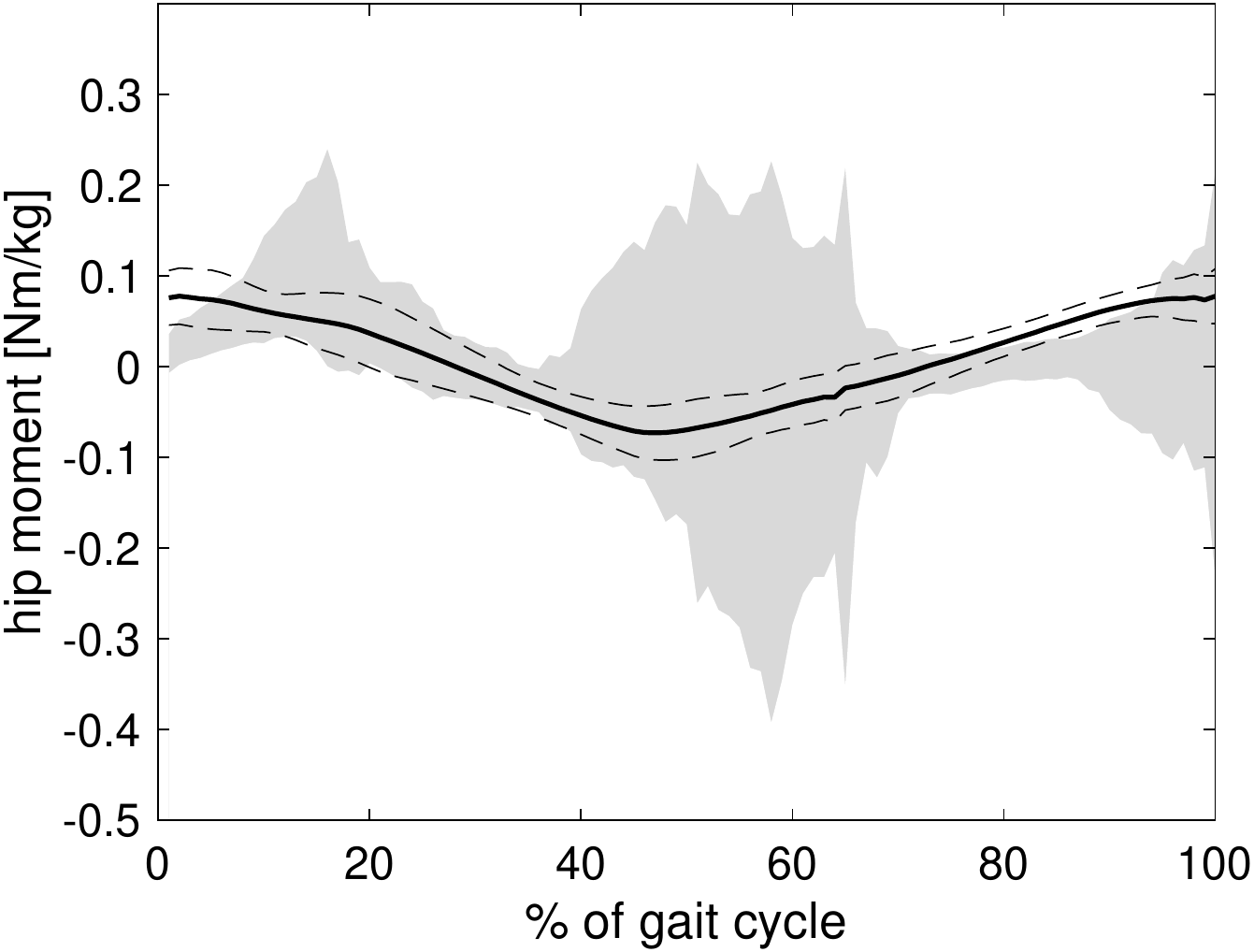}
    \end{minipage}
    \begin{minipage}{0.327\textwidth}
        \centering
        \includegraphics[width=0.95\textwidth]{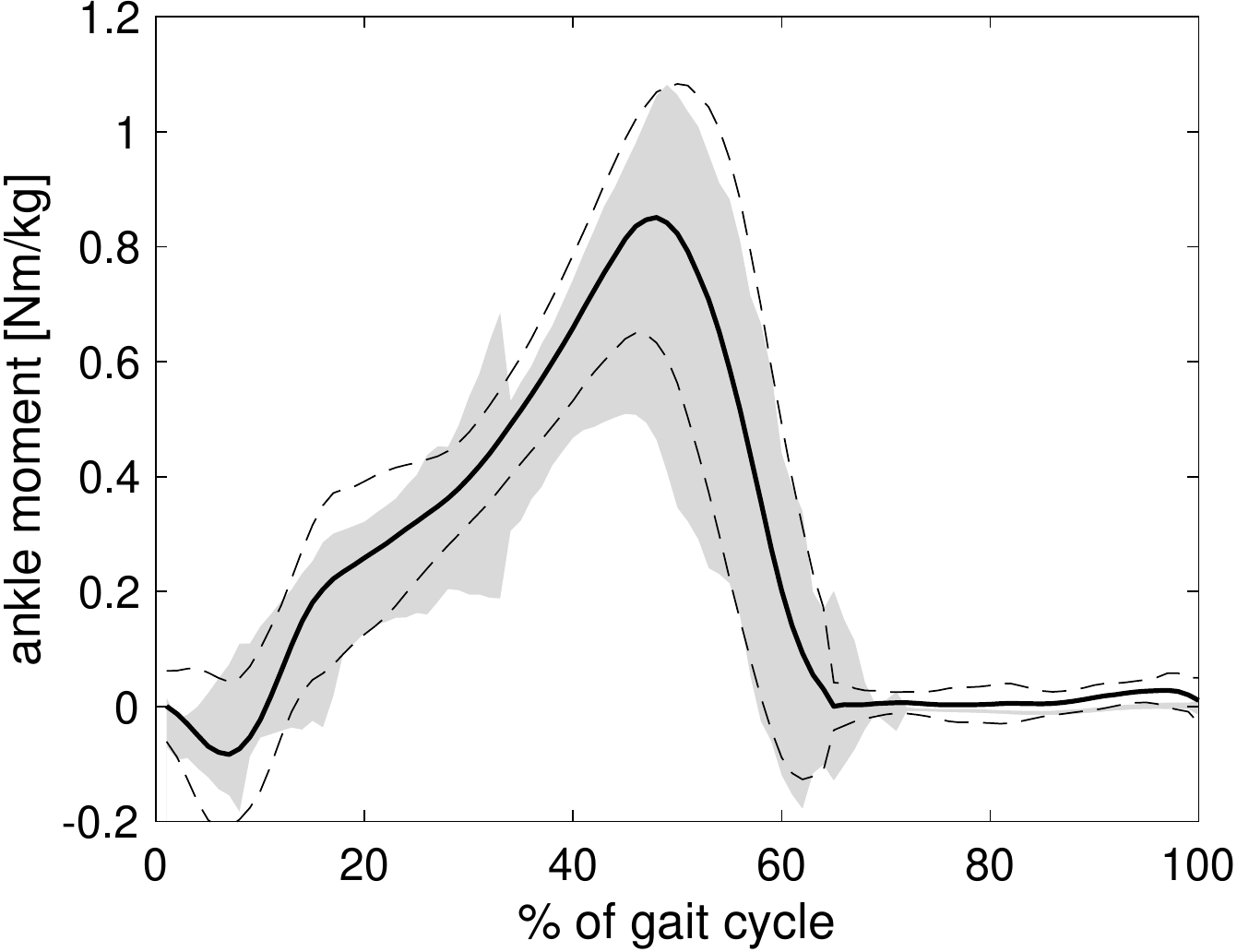}
    \end{minipage}
    \begin{minipage}{0.327\textwidth}
        \centering
        \includegraphics[width=0.95\textwidth]{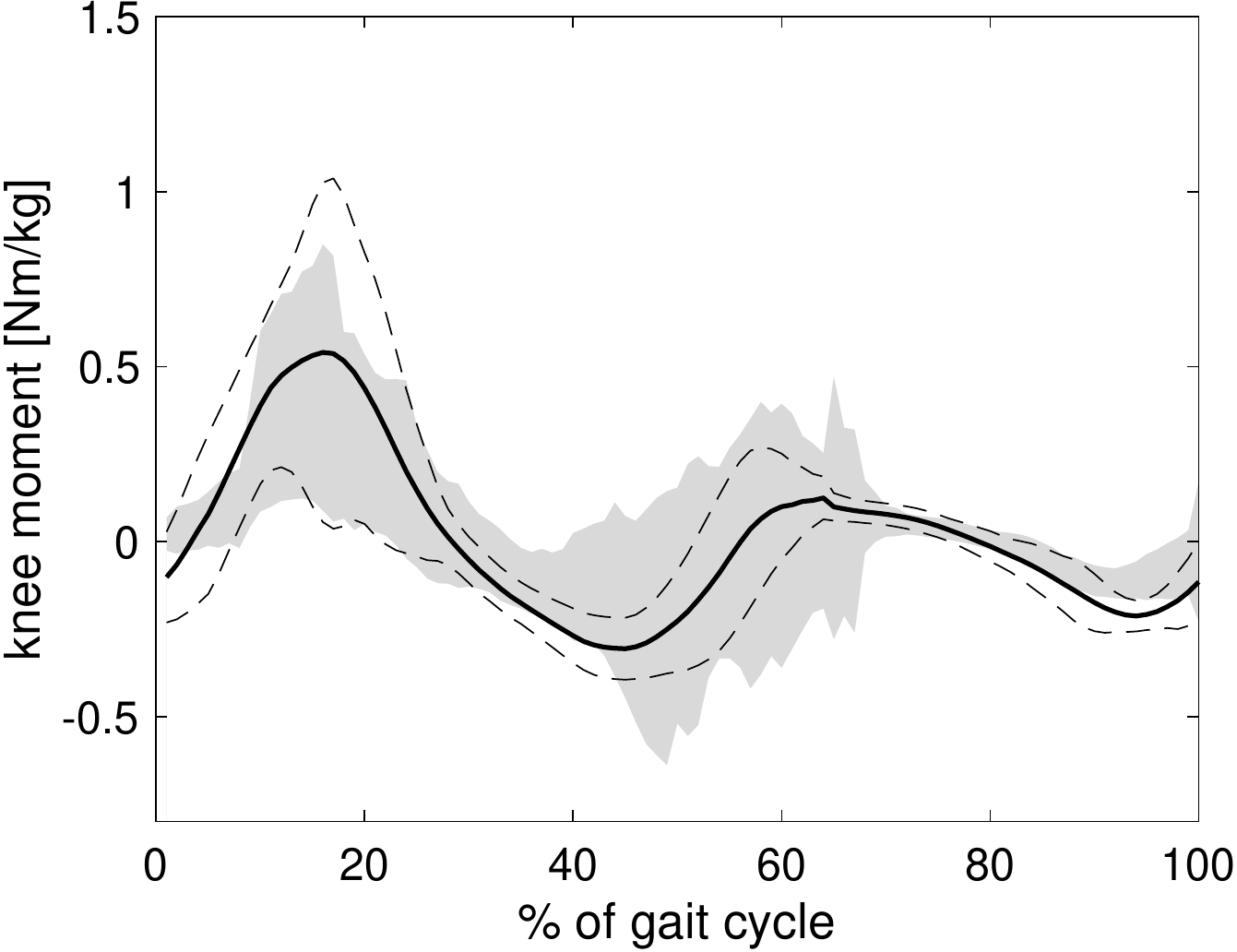}
    \end{minipage}
    \begin{minipage}{0.327\textwidth}
        \centering
        \includegraphics[width=0.95\textwidth]{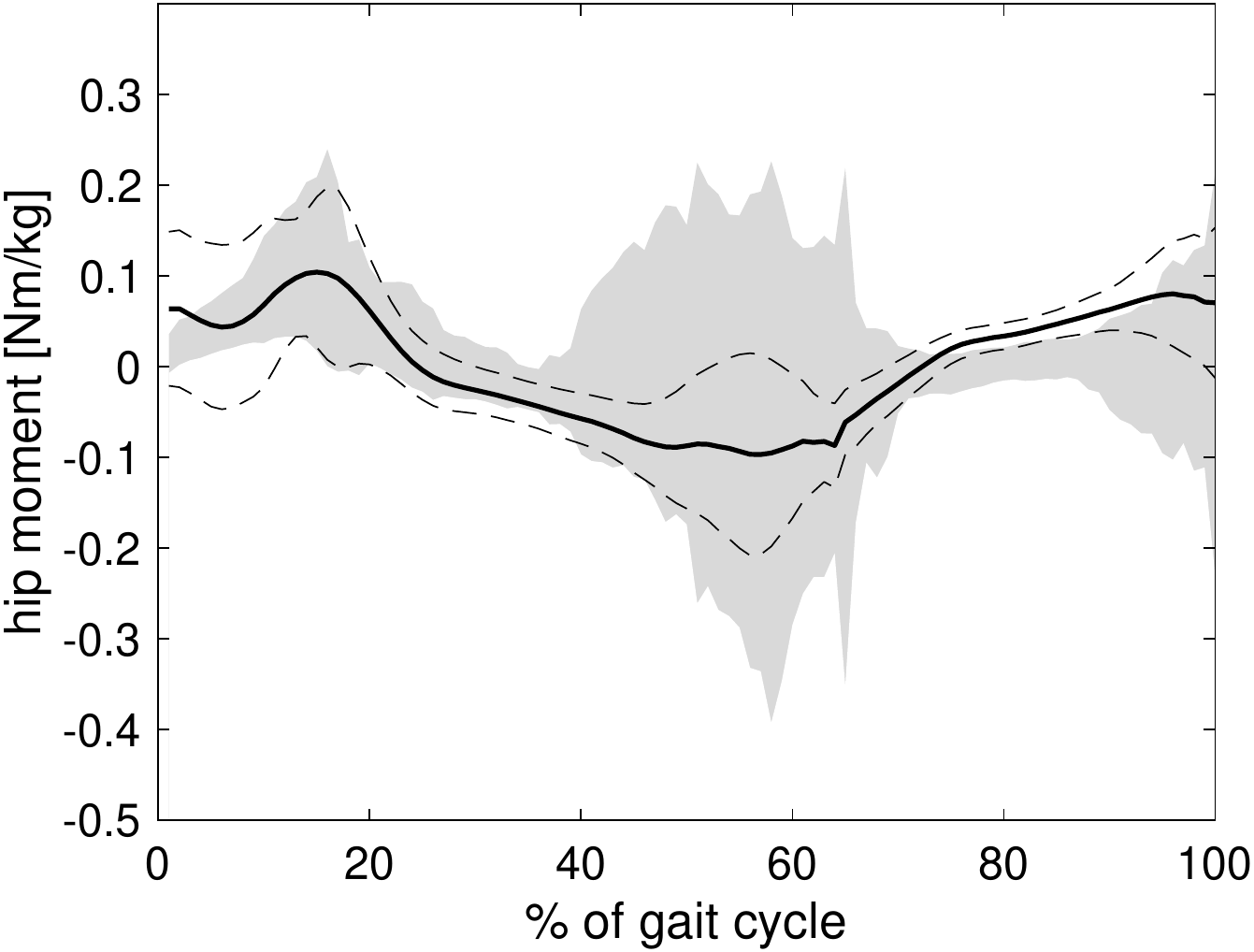}
    \end{minipage}
    
    \caption{Comparison of predicted mean curves using 10\,\% of the ground-reaction- and torque-sets. The first row shows baseline GRF/M results, the second row shows the corresponding predictions using F-net, the third row shows baseline JT results and the last row shows the corresponding results by F-net. The thick line is the mean value, the dashed lines represent standard deviations and the grey area displays the distribution of ground truth and optimized data, respectively. }
    \label{fig:ftau_curves}
\end{figure}

Figure \ref{fig:fracTau} shows the RMSE values of predicted JT and their effect in terms of RMSE of simulated motion states. The error values are illustrated for different fractions of the used ground-reaction- and torque-set. A corresponding visualization of the mean regressed GRF/M and JT curves using only 10\,\% of the ground-reaction- and torque-sets can be seen in Figure \ref{fig:ftau_curves}. A percentage of 10\,\% corresponds to one subject included in the training data, whereas the test data contains motions of 5 subjects in each validation cycle.

It can be seen, that compared to the baseline, the Dynamics Network yields stable results with significantly decreased training sets, which indicates that the cyclic training is able to compensate the lack of ground truth data by learning from the larger motion-set. With increasing number of ground-truth GRF/M samples (cf. Figure \ref{fig:fracTau} and Table \ref{tab:exp0}), F-net performs slightly better than cFI-net. This is due to a gap between the model approximation and the measured GRF/M. While F-net mainly learns ground truth GRF/M based on the included force-samples, cFI-net relies on the modelled forces to a similar extent, using the inverse-loss.

\subsection{Domain transfer}
The cyclic learning approach of the proposed method can be utilized for weakly- and unsupervised domain transfer. In the following experiment, we perform domain transfer between walking and running. For this purpose, we pre-train networks on the source motion type and apply the cyclic training modes introduced in Section \ref{sec:net} to achieve a transfer to the target motion type. During transfer learning no samples from the ground-reaction- and the torque-set are included, so that our methods operate completely independent from dynamics ground truth of the new motion type. This competence is only possible because of the novel dynamics layers. For cFI-training we use the motion-set in combination with the contact-set and minimize the corresponding loss functions alternatingly. During F-training, only the motion-set is used. For this reason, we refer to cFI-training as weakly-supervised and to F-training as unsupervised, although no ground truth data on GRF/M and JT are necessary in both cases. Table \ref{tab:trans1} lists the related results. 


\begin{table*}[]
    \centering
    \caption{Domain transfer results for a transfer between walking and running in terms of RMSE $\epsilon$ and rRMSE $\epsilon_r$ for predicted GRF/M and JT.}
    \begin{tabular}{llccccc}
    \toprule
    transfer & method & \ $\epsilon_f$ [N/kg]\ & \ $\epsilon_{r_f}$ [\%]\ & \ $\epsilon_m$ [Nm/kg]\ & \ $\epsilon_{r_m}$ [\%]\ & \ $\epsilon_\tau$ [Nm/kg]\\
    \midrule
    run & supervised & 1.388 & 23.6 & 0.091 & 21.9 & 0.041 \\
    gait to run\ \ & F-net & 3.942 & 35.6 & 0.178 & 26.1 & 0.062 \\
    gait to run\ \ & cFI-net & 1.445 & 23.6 & 0.144 & 27.0 & 0.058 \\
    gait & supervised & 0.591 & 14.4 & 0.056 & 21.2 & 0.055 \\
    run to gait\ \ & F-net & 3.579 & 37.7 & 0.144 & 34.8 & 0.217 \\
    run to gait\ \ & cFI-net & 0.685 & 13.5 & 0.076 & 33.2 & 0.153 \\
    \bottomrule
    \end{tabular}
    \label{tab:trans1}
\end{table*}

\begin{figure}
    \begin{minipage}{0.32\textwidth}
        \centering
        \includegraphics[width=0.95\textwidth]{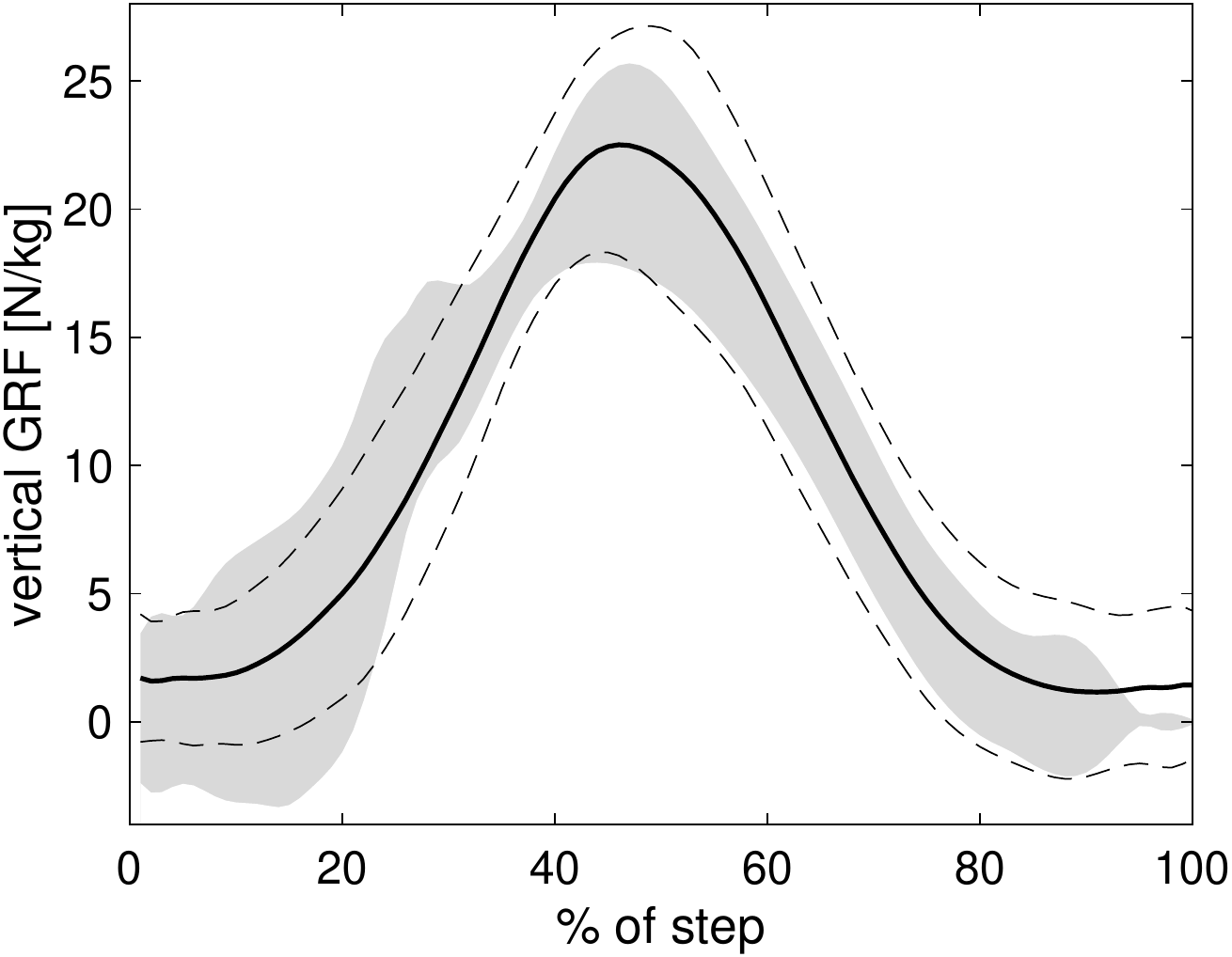}
    \end{minipage}
    \begin{minipage}{0.32\textwidth}
        \centering
        \includegraphics[width=0.95\textwidth]{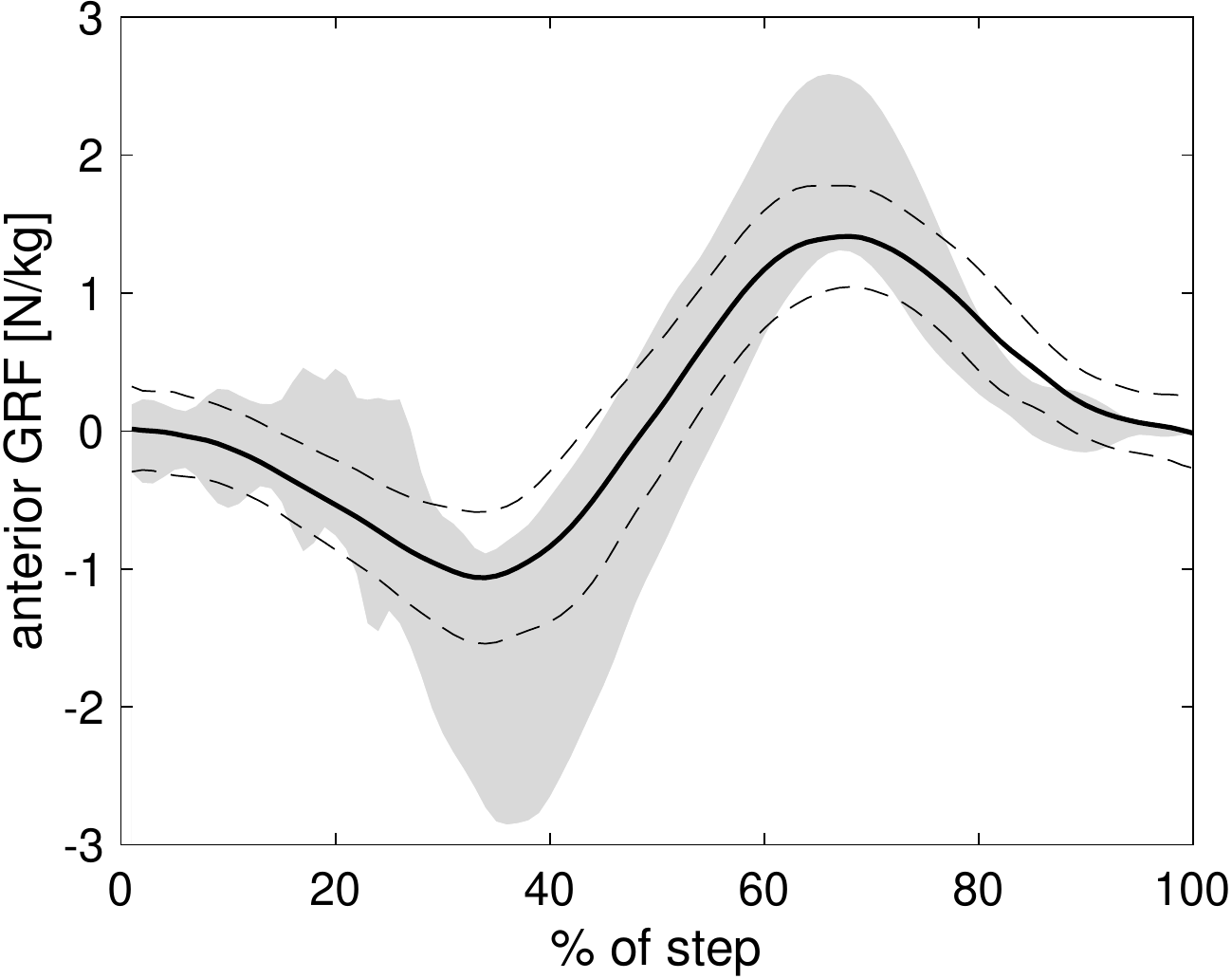}
    \end{minipage}
    \begin{minipage}{0.32\textwidth}
        \centering
        \includegraphics[width=0.95\textwidth]{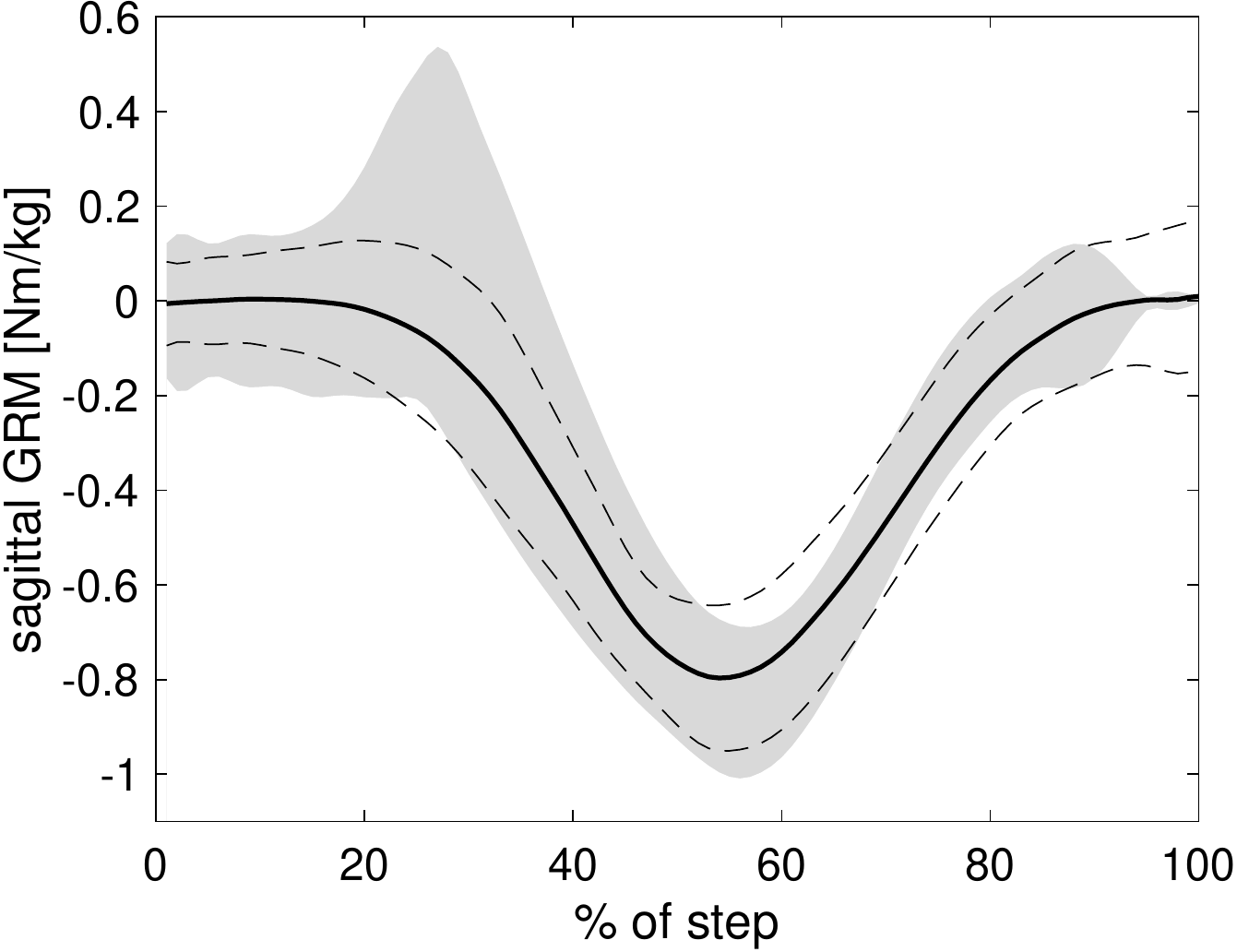}
    \end{minipage}
    \begin{minipage}{0.32\textwidth}
        \centering
        \includegraphics[width=0.95\textwidth]{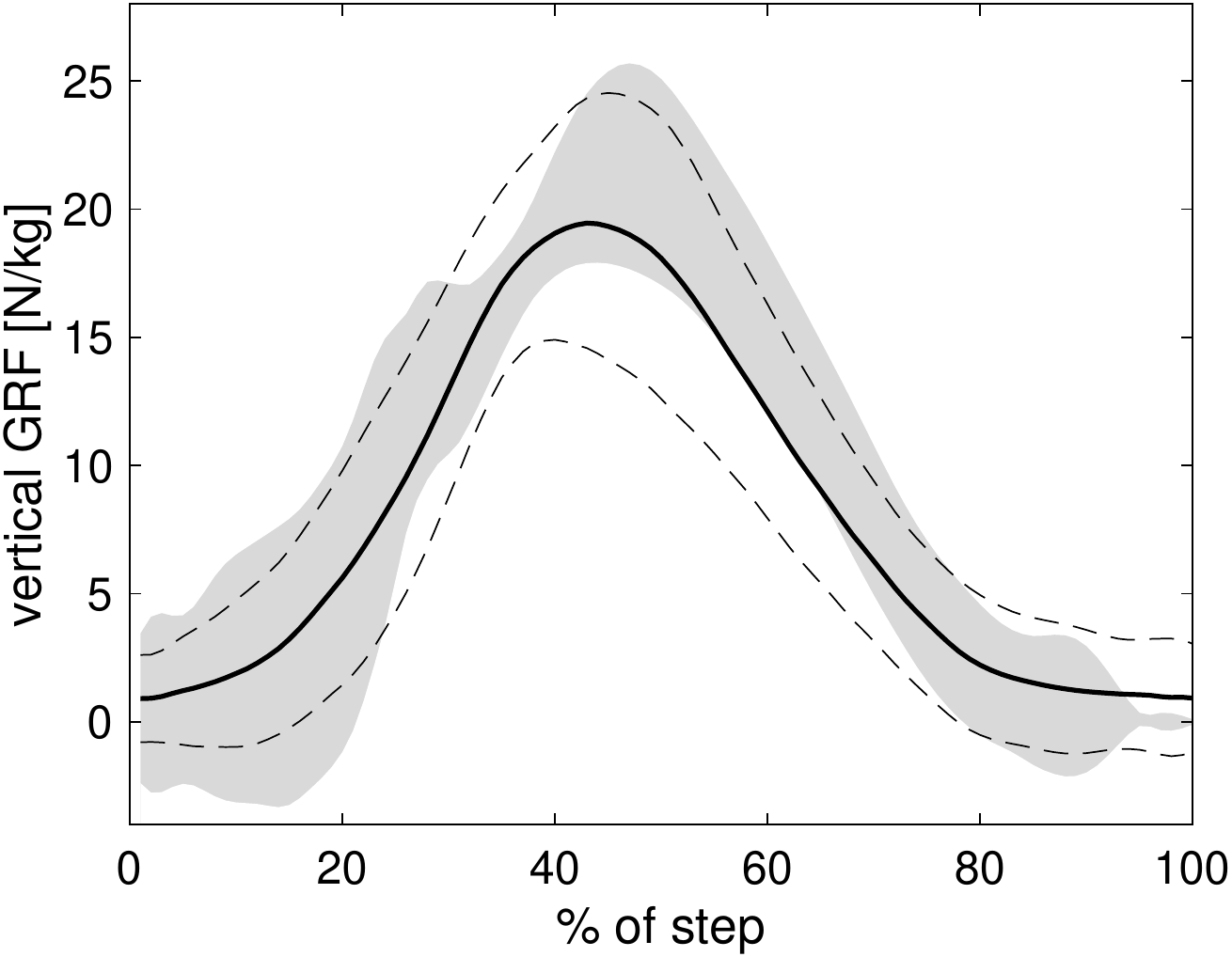}
    \end{minipage}
    \begin{minipage}{0.32\textwidth}
        \centering
        \includegraphics[width=0.95\textwidth]{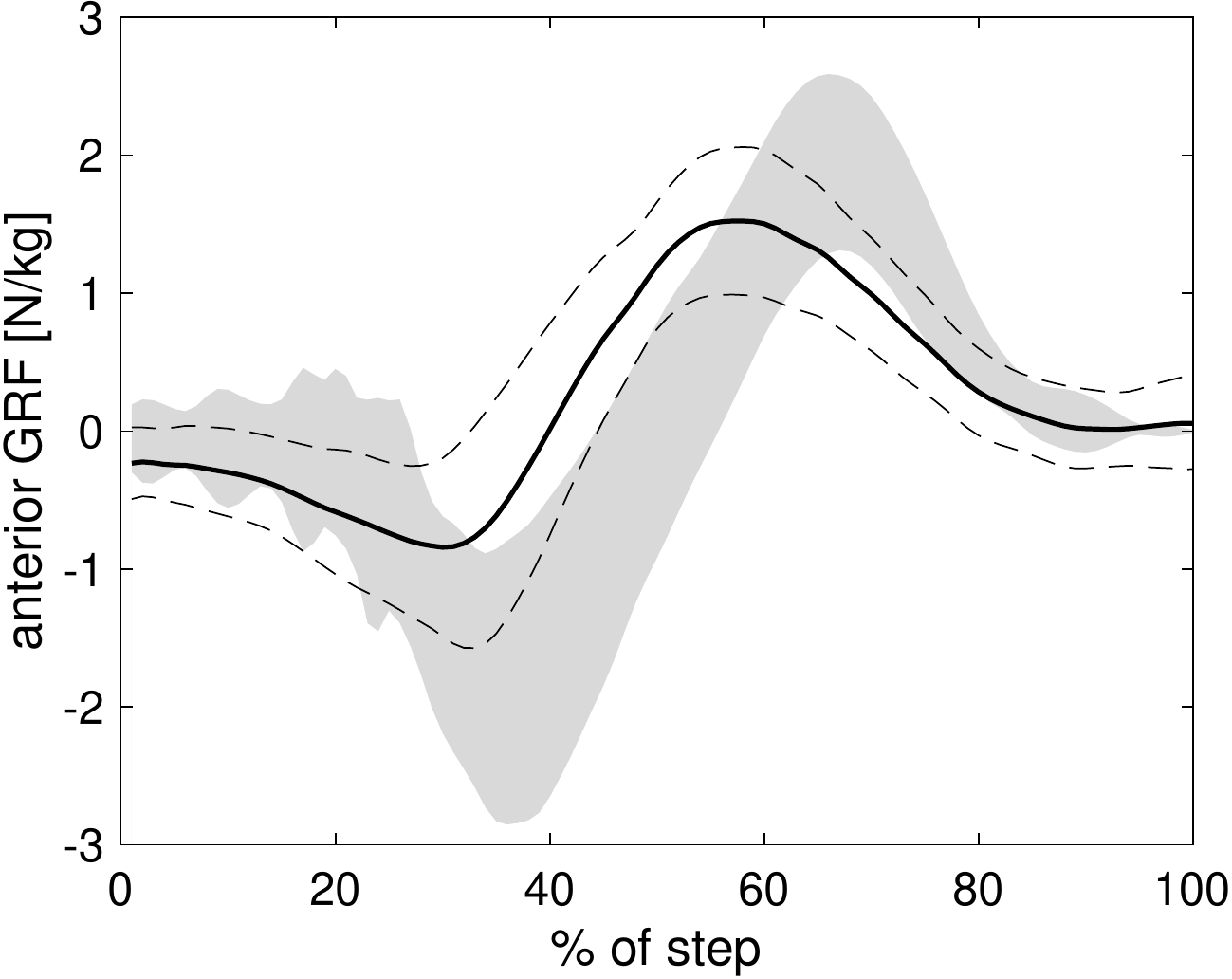}
    \end{minipage}
    \begin{minipage}{0.32\textwidth}
        \centering
        \includegraphics[width=0.95\textwidth]{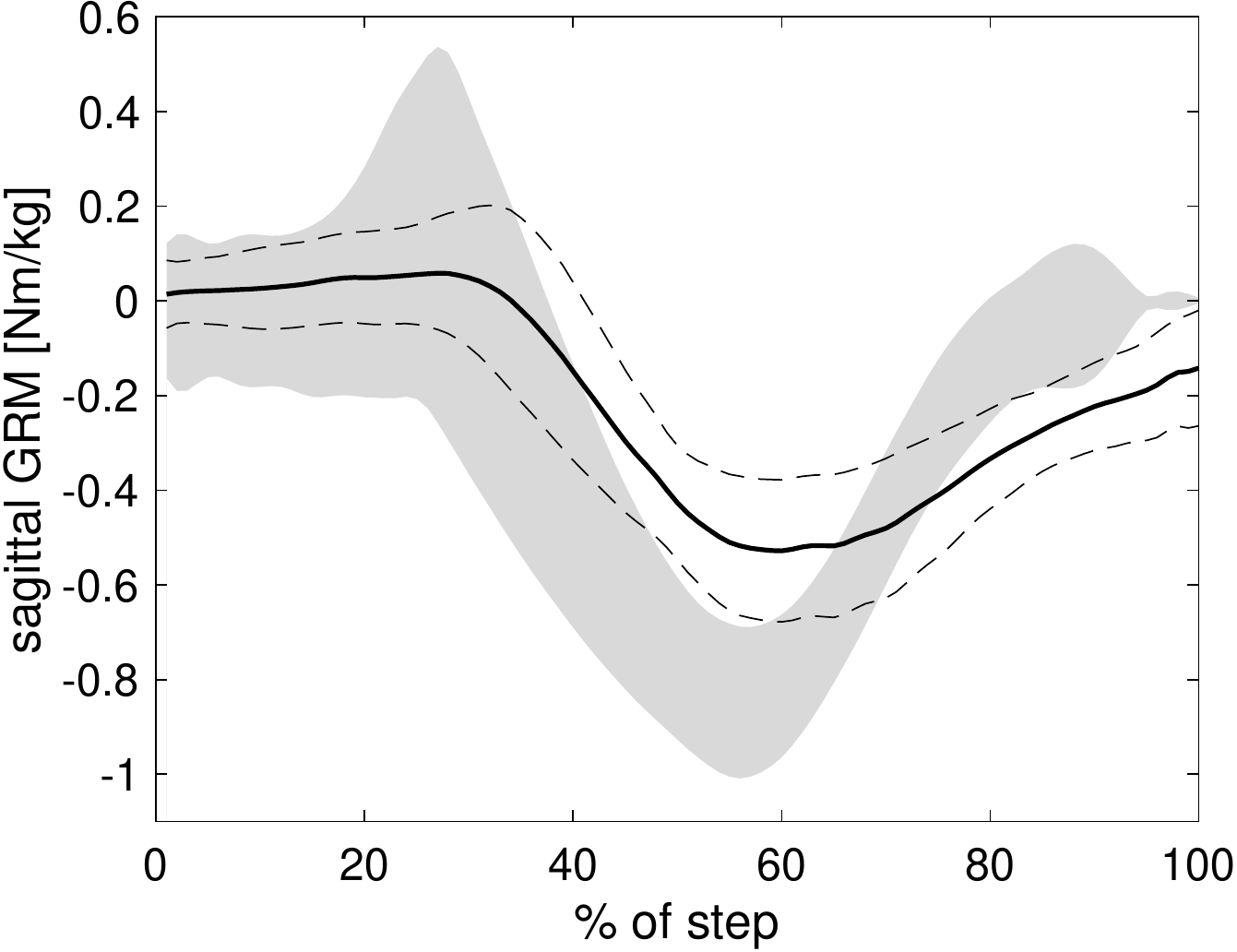}
    \end{minipage}
    \caption{Domain transfer results for the target domain \textit{running}. The top row shows baseline results, trained from scratch with supervision and the bottom row shows the results obtained using weakly-supervised transfer with cFI-net.}
    \label{fig:trans_curves}
\end{figure}

In this experiment, cFI-net clearly outperforms F-net, especially with respect to the GRF predictions. Most likely, the mutual dependency of GRF/M and JT is affecting the performance of F-net. The additional inverse layer, however, is able to solve this issue. In cFI-training the inverse layer is mainly responsible for the learning of realistic GRF/M, while the forward layer yields the matching JT. The contact-loss is necessary, since the minimization of the inverse-loss results in equally large gradients for the ground reaction of both feet. With the consideration of the contact state this overall equalisation is avoided. A visual comparison between the mean predicted GRF/M progressions for cFI-net and the supervised baseline network, is presented in Figure \ref{fig:trans_curves}.



\section{Conclusion}

This paper proposes a weakly-supervised learning approach for the inference of human dynamics. Ground reaction forces, moments and joint torques are estimated from 3D motions by means of an artificial neural network (NN), that incorporates an inverse and a forward dynamics layer to allow the minimization of a pure motion loss. The method is designed for an optimal use of currently available data of human dynamics. While there exist large public motion capture data sets, the sets that include force plate measurements are often small in comparison. In our framework, the NN estimates joint torques, as well as exterior forces and moments from motion. Together with the forward dynamics layer, that executes an integration of the equations of motion, the network performs a full cycle from motion input to the underlying forces and back to a simulated motion. This way, a pure motion loss can be minimized, allowing for weakly-supervised learning and domain transfer. The experiments show, that the proposed method achieves state-of-the-art results and performs stable, even with few ground truth samples. The domain transfer between walking and running is realized without any ground truth on contact forces and moments. This experiment demonstrates the benefit of the inverse dynamics layer. Together with a loss on the binary contact state, it reliably constrains the exterior forces. The fact that the transfer is achieved with weak-supervision offers tremendous possibilities, especially in view of recent publications in pose reconstruction \cite{Arnab_2019_CVPR}, \cite{vMarcard2018}, \cite{Sharma_2019_ICCV}, \cite{wandt2019} that make 3D human motion data readily available.\\
\textbf{Acknowledgement} Research supported by the
European Research Council (ERC-2013-PoC). The authors would like to thank all subjects who participated in data acquisition.

\clearpage
%
%
\bibliographystyle{splncs04}
\bibliography{literature}

\begin{appendix}
\section{Weight gradients of dynamics layers}
\renewcommand{\theequation}{A\arabic{equation}}
\label{app:grad}

For the forward layer, the gradients of simulated states $\bm x$ with respect to the layer input $\bm p_f$ can be calculated using sensitivity analysis: The partial differentiation of the equations of motion (Eq.~(8)) by $\bm p_f$ yields a second differential equation,
\begin{equation}
\frac{d}{dt}\frac{\partial \bm x}{\partial \bm p_f} = \frac{\partial}{\partial \bm p_f}\left( \begin{bmatrix} \bm{\dot q}\\
\bm{\mathcal{M}}^{-1}\bm{\mathcal{F}}
\end{bmatrix}\odot \bm d\right)\,,
\label{eq:gradDiff}
\end{equation}
for the gradients $\frac{\partial \bm x}{\partial \bm p}$. The integration of this differential equation supplies the desired gradients for the backpropagation through the forward layer.

Therefore, the forward pass through the forward layer includes the numerical integration of equations (8) and (\ref{eq:gradDiff}) with storage of the resulting gradients. During the backward pass the gradient of the loss is propagated by multiplication with the stored gradients according to the chain rule:
\begin{equation}
\frac{\partial L_\mathrm{forward}}{\partial \bm p_f} = \sum_{t=1}^n\frac{\partial L_\mathrm{forward}}{\partial \bm x^{sim}_t}\frac{\partial \bm x^{sim}_t}{\partial \bm p_f}\,.
\end{equation}
The calculation of gradients for the inverse layer is straight forward, since there is no dependency between the resulting residual forces and moments of different time frames. Given the relevant input $\bm p_i=\bm F_{c_t}$, the gradients can be calculated by
\begin{equation}
\frac{\partial L_\mathrm{inverse}}{\partial \bm p_i} = \frac{2}{n}\sum_{t=1}^n \left(\bm F_{\mathrm{res}_t}\frac{\partial \bm F_{\mathrm{res}_t}}{\partial \bm F_{c_t}}+ \bm M_{\mathrm{res}_t}\frac{\partial \bm M_{\mathrm{res}_t}}{\partial \bm F_{c_t}}\right)\,.
\end{equation}
\newpage
\section{Demographic information}
The recorded data set encompasses 195 walking and 75 running sequences executed by 22 healthy subjects. Demographic information can be found in Table \ref{tab:demographic}. All subjects volunteered to participate in the study and signed an informed consent form. The study is part of the ‘‘Individualized Implant Placement’’ project funded by the European Research Council (ERC-2013-PoC) and was approved by the ethics commission of the Hannover Medical School (MHH). In order to increase and balance our data set, we augment by mirroring the kinematics and dynamics at the sagittal plane \footnote{The sagittal plane is spanned by the vertical axis and the direction of movement.}. 
\renewcommand{\thetable}{B\arabic{table}}
\label{tab:demographic}
\begin{table}
	\caption{Demographic table of participating subjects.}
	\label{tab:demographic}
	\centering
	\begin{tabular}{c c c c c}
		\toprule
		subject ID & gender & height & weight & BMI \\
		\midrule
		1 & m & 1.78 & 93.5 & 30\\
		2 & m & 1.94 & 88.8 & 24\\
		3 & m & 1.86 & 68.3 & 20\\
		4 & f & 1.71 & 66.6 & 23\\
		5 & m & 1.80 & 68.3 & 21\\
		6 & f & 1.73 & 55.7 & 19\\
		7 & f & 1.69 & 65.5 & 23\\
		8 & m & 1.71 & 61.8 & 21\\
		9 & m & 1.81 & 67.9 & 21\\
		10 & m & 1.88 & 74.4 & 21\\
		11 & m & 1.81 & 79.3 & 24\\
		12 & m & 1.85 & 74.5 & 22\\
		13 & m & 1.67 & 83.8 & 30\\
		14 & m & 1.85 & 95.8 & 28\\
		15 & m & 1.84 & 68.8 & 20\\
		16 & m & 1.75 & 81.4 & 27\\
		17 & m & 1.72 & 79.4 & 27\\
		18 & f & 1.70 & 68.0 & 24\\
		19 & m & 1.80 & 72.4 & 22\\
		20 & f & 1.74 & 70.5 & 23\\
		21 & m & 1.80 & 83.5 & 26\\
		22 & m & 1.79 & 69.9 & 22\\
		\bottomrule
	\end{tabular}
\end{table}

\section{Application to CMU data}
\begin{figure}
	\centering
	\includegraphics[width=0.5\textwidth]{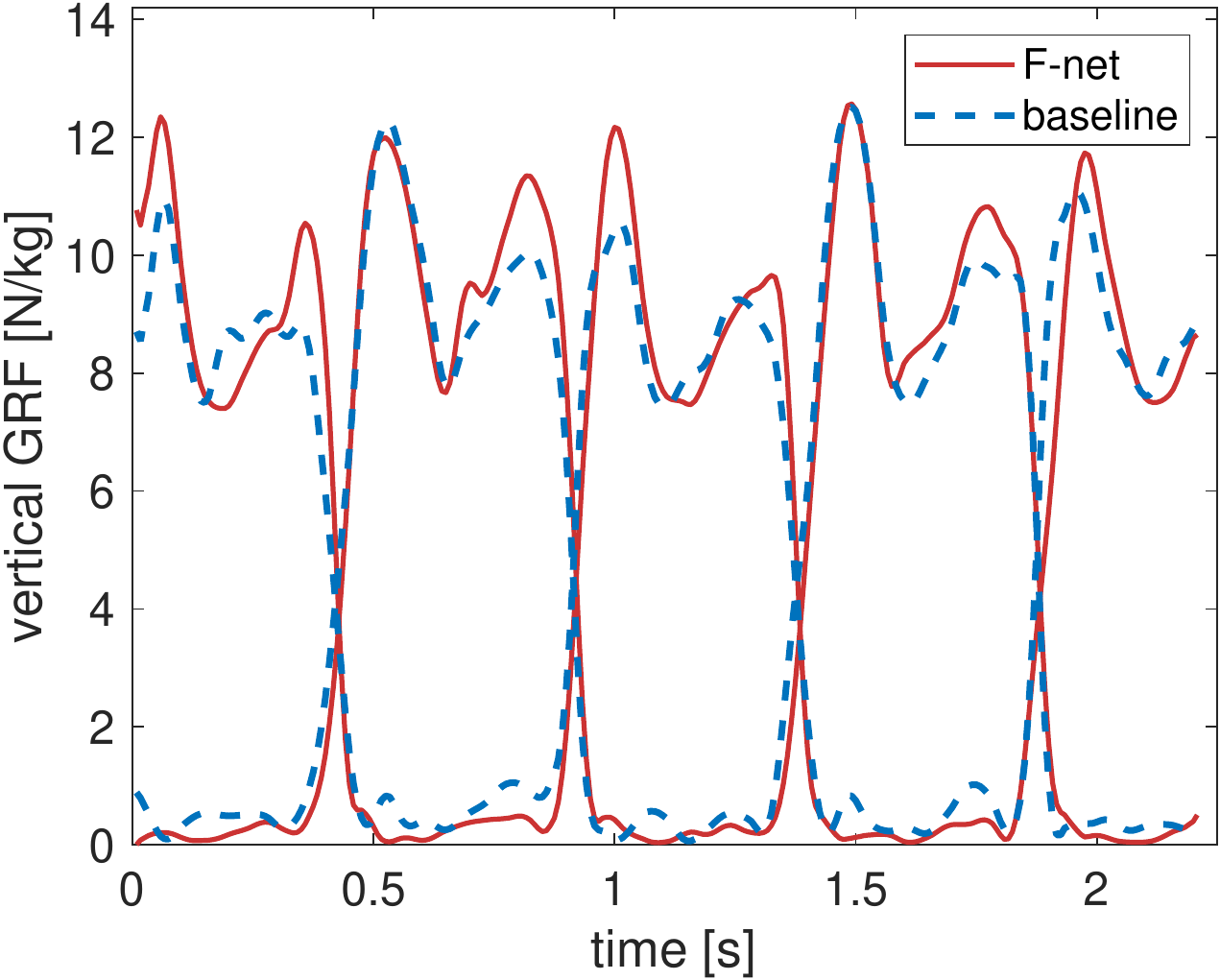}
	\caption{Regressed vertical GRF for an example sequence taken from the CMU data base.}
	\label{fig:cmu}
\end{figure}

In order to test the generalizability to a different 3D motion data set, we apply our method to walking sequences, taken from the CMU data base \cite{cmumocap}. Figure \ref{fig:cmu} shows an exemplary comparison between the baseline method and F-net. Since the baseline network can only be trained in a supervised manner, it is exclusively trained on our own data set. F-net is additionally trained on CMU-samples using the forward-loss. The result indicates, that the additional cyclic training helps to bridge the domain gap between the two sets: In contrast to the baseline, F-net yields symmetric vertical GRF for both feet and the forces stay closer to zero during frames without contact.

\section{Noise experiment}
Zero mean gaussian noise with standard deviation $\sigma$ is added to the joint angles of the motion training set and the test sequences. The results are presented in Table \ref{tab:noise}. In the case of cFI-net we simulate noisy contact detections in addition to the angle noise. For this purpose, 10\,\% of the contact labels are randomly chosen and switched with a probability of 50\,\%.

Angle noise of $\sigma=0.3$ deg only marginally effects the performance of both networks. Overall, F-net is slightly more robust against noise than cFI-net, in this experiment. The inverse layer matches GRF/M to the noisy motion, resulting in a larger difference between inferred forces and ground truth. In contrast to that, F-net primarily learns GRF/M from the ground truth set. Concerning the JT, the robustness of both methods is comparable. During training, the JT output of both networks is controlled by the MSE and the forward-loss in a similar way.
\begin{table*}[]
	\centering
	\caption{Influence of angle noise: RMSE $\epsilon$ and rRMSE $\epsilon_r$ of GRF/M and JT regression results for the gait data set with noisy motion training set and noisy test sequences.}
	\begin{tabular}{lcccccc}
		\toprule
		method & $\sigma$ [deg]&\ $\epsilon_f$ [N/kg]\ & \ $\epsilon_{r_f}$ [\%]\ & \ $\epsilon_m$ [Nm/kg]\ & \ $\epsilon_{r_m}$ [\%]\ & \ $\epsilon_\tau$ [Nm/kg]\\
		\midrule
		F-net& 0.3 & 0.626 & 14.9 & 0.058 & 22.4 & 0.062 \\
		& 0.6 & 0.832 & 16.8 & 0.061 & 22.7 & 0.068 \\
		& 1.1 & 0.898 & 17.8 & 0.066 & 24.3 & 0.073 \\
		& 2.3 & 1.096 & 19.7 & 0.073 & 25.4 & 0.092 \\
		\midrule
		cFI-net& 0.3 & 0.674 & 13.8 & 0.062 & 23.4 & 0.060 \\
		& 0.6 & 0.908 & 16.7 & 0.088 & 26.4 & 0.076 \\
		& 1.1 & 0.998 & 17.5 & 0.082 & 25.9 & 0.087 \\
		& 2.3 & 1.143 & 19.5 & 0.079 & 25.7 & 0.096 \\
		\bottomrule
	\end{tabular}
	\label{tab:noise}
\end{table*}

\end{appendix}
\end{document}